%% file: main.tex
\documentclass[a4paper]{article}

\usepackage[english]{babel}
\usepackage[utf8]{inputenc}
\usepackage{amsmath}
\usepackage{graphicx}
\usepackage{amssymb}

\usepackage[margin=1in]{geometry}

\usepackage{hyperref}
\usepackage{bm}
\usepackage{url}
\usepackage{wrapfig}
\usepackage[usenames,dvipsnames,svgnames,table]{xcolor}
\usepackage{soul}	
\usepackage{fancyvrb}
\usepackage{framed}

\newtheorem{theorem}{Theorem}

\newif\ifcomment
\commenttrue      

\ifcomment
\newcommand{\qirong}[1]{{\color{blue}{\bf\sf [Qirong: #1]}}} 
\newcommand{\daiwei}[1]{{\color{purple}{\bf\sf [Daiwei: #1]}}}

\else
\newcommand{\qirong}[1]{{\color{blue}{}}}
\newcommand{\abhi}[1]{{\color{magenta}{}}}
\newcommand{\daiwei}[1]{{\color{purple}{}}}
\newcommand{\jinliang}[1]{{\color{green}{}}}
\fi

\newcommand{\xv}{x}
\newcommand{\yv}{y}
\newcommand{\av}{A}
\newcommand{\obj}{\mathcal{L}}
\newcommand{\uv}{u}

\def\Var{{\rm Var}\,}

\newcommand{\mb}{\mathbf}

\newcommand{\xvec}{\mathbf{x}}

\newcommand{\zvec}{\mathbf{z}}
\newcommand{\RR}{\mathbb{R}}
\newcommand{\EE}{\mathbb{E}}

\usepackage{xspace}
\makeatletter
\DeclareRobustCommand\onedot{\futurelet\@let@token\@onedot}
\def\@onedot{\ifx\@let@token.\else.\null\fi\xspace}

\title{Strategies and Principles of Distributed Machine Learning on Big Data}

\author{Eric P. Xing, Qirong Ho, Pengtao Xie, Wei Dai \\ School of Computer Science, Carnegie Mellon University}

\begin{document}

\maketitle

\begin{abstract}
The rise of Big Data has led to new demands for Machine Learning (ML) systems to learn complex models with millions to billions of parameters, that promise adequate capacity to digest massive datasets and offer powerful predictive analytics (such as high-dimensional latent features, intermediate representations, and decision functions) thereupon. In order to run ML algorithms at such scales, on a distributed cluster with 10s to 1000s of machines, it is often the case that significant engineering efforts are required --- and one might fairly ask if such engineering truly falls within the domain of ML research or not. Taking the view that Big ML systems can benefit greatly from ML-rooted statistical and algorithmic insights --- and that ML researchers should therefore not shy away from such systems design --- we discuss a series of principles and strategies distilled from our recent efforts on industrial-scale ML solutions. These principles and strategies span a continuum from application, to engineering, and to theoretical research and development of Big ML systems and architectures, with the goal of understanding how to make them efficient, generally-applicable, and supported with convergence and scaling guarantees. They concern four key questions which traditionally receive little attention in ML research: How to distribute an ML program over a cluster? How to bridge ML computation with inter-machine communication? How to perform such communication? What should be communicated between machines? By exposing underlying statistical and algorithmic characteristics unique to ML programs but not typically seen in traditional computer programs, and by dissecting successful cases to reveal how we have harnessed these principles to design and develop both high-performance distributed ML software as well as general-purpose ML frameworks, we present opportunities for ML researchers and practitioners to further shape and grow the area that lies between ML and systems.
\end{abstract}

\input{intro}

\input{background}
\input{principles}
\input{conclusion}

\bibliographystyle{abbrv}
\bibliography{literature}

\end{document}

%% file: intro.tex
\section{Introduction}

Machine Learning (ML) has become a primary mechanism for distilling structured information and knowledge from raw data, turning them into automatic predictions and actionable hypotheses for diverse applications, such as analyzing social networks~\cite{airoldi2008mixed}, reasoning about customer behaviors~\cite{ahmed2011unified}, interpreting texts, images and videos~\cite{zhao2014quasi}, identifying disease and treatment paths~\cite{lee2012leveraging}, driving vehicles without the need for a human~\cite{ventures2006stanley}, and tracking anomalous activity for cybersecurity~\cite{chandola2009anomaly}, amongst others. The majority of ML applications are supported by a moderate number of families of well-developed {\it ML approaches}, each of which embodies a continuum of technical elements from model design, to algorithmic innovation, and even to perfection of the software implementation, and which attracts ever-growing novel contributions from the research and development community. Modern examples of such approaches include Graphical Models~\cite{wainwright2008graphical,koller2009probabilistic,cmu10708}, Regularized Bayesian models~\cite{zhu2009maximum,zhu2009medlda,zhu2014bayesian}, Nonparametric Bayesian models~\cite{ghahramani2005infinite,teh2006hierarchical}, Sparse Structured models~\cite{yuan2006model,kim2012tree}, Large-margin methods~\cite{burges1998tutorial,roller2004max}, Deep learning~\cite{hinton2012deep,krizhevsky2012imagenet}, Matrix Factorization~\cite{lee1999learning,mnih2007probabilistic}, Sparse Coding~\cite{olshausen1997sparse,lee2006efficient}, and Latent Space Modeling~\cite{airoldi2008mixed,zheng2015model}. A common ML practice that ensures mathematical soundness and outcome reproducibility is for practitioners and researchers to write an {\it ML program} (using any generic high-level programming language) for an application-specific {\it instance} of a particular ML approach (e.g. semantic interpretation of images via a deep learning model such as a convolution neural network). Ideally this program is expected to execute quickly and accurately on a variety of hardware and cloud infrastructure: laptops, server machines, GPUs, cloud compute and virtual machines, distributed network storage, Ethernet and Infiniband networking, just to name a few. Thus, the program is {\it hardware-agnostic} but {\it ML-explicit} (i.e., following the same mathematical principle when trained on data and attaining the same result regardless of hardware choices.).

With the advancements in sensory, digital storage, and Internet communication technologies, conventional ML research and development --- which excels in model, algorithm, and theory innovations --- are now challenged by the growing prevalence of Big Data collections, such as hundreds of hours of video uploaded to video-sharing sites every minute\footnote{\url{https://www.youtube.com/yt/press/statistics.html}}, or petabytes of social media on billion-plus-user social networks\footnote{\url{https://code.facebook.com/posts/229861827208629/scaling-the-facebook-data-warehouse-to-300-pb/}}. The rise of Big Data is also being accompanied by an increasing appetite for higher-dimensional and more complex ML models with billions to trillions of {\it parameters}, in order to support the ever-increasing complexity of data, or to obtain still higher predictive accuracy (e.g. for better customer service and medical diagnosis) and support more intelligent tasks (e.g. driver-less vehicles and semantic interpretation of video data)~\cite{yuan2015lightlda,coates2013deep}. Training such Big ML Models over such Big Data is beyond the storage and computation capabilities of a single machine, and this gap has inspired a growing body of recent work on {\it distributed ML}, where ML programs are executed across research clusters, data centers and cloud providers with 10s to 1000s of machines. Given $P$ machines instead of one machine, one would expect a nearly $P$-fold speedup in the time taken by a distributed ML program to complete, in the sense of attaining a mathematically equivalent or comparable solution to that produced by a single machine; yet, the reported speedup often falls far below this mark --- for example, even recent state-of-the-art implementations of topic models~\cite{ahmed2012scalable} (a popular method for text analysis) cannot achieve $2\times$ speedup with $4\times$ machines, because of mathematical incorrectness in the implementation (as shown in~\cite{zheng2015model}), while deep learning on MapReduce-like systems such as Spark has yet to achieve $5\times$ speedup with $10\times$ machines~\cite{moritz2015sparknet}. Solving this {\it scalability challenge} is therefore a major goal of distributed ML research, in order to reduce the {\it capital and operational cost} of running Big ML applications.

Given the iterative-convergent nature of most --- if not all --- major ML algorithms powering contemporary large scale applications, at a first glance, one might naturally identify two possible avenues toward scalability: {\it faster convergence} as measured by iteration number (also known as {\it convergence rate} in the ML community), and {\it faster per-iteration time} as measured by the actual speed at which the system executes an iteration (also known as {\it throughput} in the system community). Indeed, a major current focus by many distributed ML researchers
is on algorithmic correctness as well as faster {\it convergence rates} over a wide spectrum of ML approaches~\cite{agarwal2011distributed,hogwild} However, many of the ``accelerated" algorithms from this line of research face difficulties in making their way to industry-grade implementations, because of their idealized assumptions on the system --- for example, the assumption that networks are infinitely fast (i.e. zero synchronization cost), or the assumption that all machines make algorithm progress at the same rate (implying no background tasks and only a single user of the cluster, which are unrealistic expectations for real-world research and production clusters shared by many users). On the other hand, systems researchers focus on high {\it iteration throughput} (more iterations per second) and fault-recovery guarantees, but may choose to assume that the ML algorithm will work correctly under non-ideal execution models (such as fully asynchronous execution), or that it can be rewritten easily under a given abstraction (such as MapReduce or Vertex Programming)~\cite{dean2008mapreduce,gonzalez2012powergraph,zaharia2012resilient}. In both ML and systems research, issues from the other side can become oversimplified, which may in turn obscure new opportunities to reduce the capital cost of distributed ML. In this paper, we propose a strategy that combines ML-centric and system-centric thinking, and in which the nuances of both ML algorithms (mathematical properties) and systems hardware (physical properties) are brought together to allow insights and designs from both ends to work in concert and amplify each other.

Many of the existing general-purpose Big Data software platforms present a unique tradeoff among correctness, speed of execution, and ease-of-programmability for ML applications. For example, dataflow systems such as Hadoop and Spark~\cite{zaharia2012resilient} are built on a MapReduce-like abstraction~\cite{dean2008mapreduce} and provide an easy-to-use programming interface, but have paid less attention to ML properties such as error tolerance, fine-grained scheduling of computation and communication to speed up ML programs --- as a result, they offer correct ML program execution and easy programming, but are slower than ML-specialized platforms~\cite{xing2015petuum,li2014scaling}. This (relative) lack of speed can be partly attributed to the bulk synchronous parallel (BSP) {\it synchronization model} used in Hadoop and Spark, where machines assigned to a group of tasks must wait at a barrier for the slowest machine to finish, before proceeding with the next group of tasks (e.g. all mappers must finish before the reducers can start)~\cite{ho2013more}. Another example are the graph-centric platforms such as GraphLab and Pregel, which rely on a graph-based ``vertex programming" abstraction that opens up new opportunities for ML program partitioning, computation scheduling, and flexible consistency control --- hence, they are usually correct and fast for ML. However, ML programs are not usually conceived as vertex programs (instead, they are mathematically formulated as iterative-convergent fixed-point equations), and it requires non-trivial effort to rewrite them as such. In a few cases, the graph abstraction may lead to incorrect execution or suboptimal execution speed~\cite{kumar2014fugue,lee2014model}. Of recent note is the parameter server paradigm~\cite{ho2013more,dai2015high,wei2015managed,ahmed2012scalable,li2014scaling}, which provides a ``design template" or philosophy for writing distributed ML programs from the ground-up, but is not a programmable platform or work partitioning system in the same sense as Hadoop, Spark, GraphLab and Pregel. Taking into account the common ML practice of writing ML programs for application-specific instances, a usable software platform for ML practitioners could instead offer two utilities: (1) a ready-to-run set of {\it ML workhorse implementations} --- such as stochastic proximal descent algorithms~\cite{bottou2010large, mspg}, coordinate descent algorithms~\cite{fercoq2013accelerated}, Markov Chain Monte Carlo algorithms~\cite{gilks2005markov} --- that can be re-used across different ML algorithm families. In turn, these workhorse implementations are supported by (2) an {\it ML Distributed Cluster Operating System}, which partitions and executes these workhorses across a wide variety of hardware. Such a software platform not only realizes the capital cost reductions obtained through distributed ML research, but even complements it by reducing the {\it human cost} (scientist- and engineer-hours) of Big ML applications, through easier-to-use programming libraries and cluster management interfaces.

With the growing need to enable data-driven knowledge distillation, decision making, and perpetual learning --- which are representative hallmarks of the vision for machine intelligence --- in the coming years, the major form of computing workloads on Big Data is likely to undergo a rapid shift from database-style operations for deterministic storage, indexing, and queries, to ML-style operations such as probabilistic inference, constrained optimization, and geometric transformation. To best fulfill these computing tasks, which must perform a large number of passes over the data and solve a high-dimensional mathematical program, there is a need to revisit the principles and strategies in traditional system architectures, and explore new designs that optimally balance correctness, speed, programmability, and deployability.
A key insight necessary for guiding such explorations is an understanding that ML programs are {\it optimization-centric}, and frequently admit {\it iterative-convergent} algorithmic solutions rather than one-step or closed form solutions. Furthermore, ML programs are characterized by three properties: (1) {\it error tolerance}, which makes ML programs robust against limited errors in intermediate calculations; (2) {\it dynamic structural dependencies}, where the changing correlations between model parameters must be accounted for in order to achieve efficient, near-linear parallel speedup; (3) {\it non-uniform convergence}, where each of the billions (or trillions) of ML parameters can converge at vastly different iteration numbers (typically, some parameters will converge in 2-3 iterations, while others take hundreds). These properties can be contrasted with traditional programs (such as sorting and database queries), which are {\it transaction-centric} and only guaranteed to execute correctly if every step is performed with {\it atomic correctness}~\cite{dean2008mapreduce,zaharia2012resilient}. In this paper, we shall derive unique design principles for distributed ML systems based on these properties; these design principles strike a more effective balance between ML correctness, speed and programmability (while remaining generally applicable to almost all ML programs), and are organized into four upcoming sections: (I) How to distribute ML programs; (II) How to bridge ML computation and communication; (III) How to communicate; (IV) What to communicate. Before delving into the principles, let us first review some necessary background information about iterative-convergent ML algorithms.

%% file: background.tex

\section{Background: Iterative-Convergent ML Algorithms}
\label{sec:background}

With a few exceptions, almost all ML programs can be viewed as {\it optimization-centric} programs that adhere to a general mathematical form:
\begin{align}
&\max_\av \obj(\mathbf{x},\av) \quad \text{OR} \quad \min_\av \obj(\mathbf{x},\av), \label{eq:ml_form}\\
\text{where}\quad &\obj(\mathbf{x},\av) = f(\{x_i,y_i\}_{i=1}^N ; \av) + r(\av). \notag
\end{align}
In essence, an ML program tries to fit $N$ data samples (which may be {\it labeled} or {\it unlabeled}, depending on the real-world application being considered), represented by $\mathbf{x} \equiv \{x_i,y_i\}_{i=1}^N$ (where $y_i$ is present only for {\it labeled} data samples), to a model represented by $\av$. This fitting is performed by optimizing (maximizing or minimizing) an overall objective function $\obj$, composed of two parts: a {\it loss function} $f$ that describes how data should fit the model, and a {\it structure-inducing} function $r$ that incorporates domain-specific knowledge about the intended application, by placing constraints or penalties on the values $\theta$ can take.

The apparent simplicity of Eq.~\ref{eq:ml_form} belies the potentially complex structure of the functions $f,r$, and the potentially massive size of the data $\mathbf{x}$ and model $\av$. Furthermore, ML algorithm families are often identified by their unique characteristics on $f,r,\mathbf{x},\av$. For example, a typical {\it deep learning} model for image classification, such as~\cite{krizhevsky2012imagenet}, will contain 10s of millions through billions of matrix-shaped model parameters in $\av$, while the loss function $f$ exhibits a deep recursive structure $f() = f_1(f_2(f_3(\dots) + \dots) + \dots)$ that learns a hierarchical representation of images similar to the human visual cortex. {\it Structured sparse regression} models~\cite{lee2012leveraging} for identifying genetic disease markers may use overlapping structure-inducing functions $r() = r_1(\av_a) + r_2(\av_b) + r_3(\av_c) + \dots$, where $\av_a,\av_b,\av_c$ are overlapping subsets of $\av$, in order to respect the intricate process of chromosomal recombination. {\it Graphical models}, particularly Topic models, are routinely deployed on billions of documents $\mathbf{x}$ --- i.e. $N\ge 10^9$, a volume that is easily generated by social media such as Facebook and Twitter --- and can involve up to trillions of parameters $\theta$ in order to capture rich semantic concepts over so much data~\cite{yuan2015lightlda}.

Apart from specifying Eq.~\ref{eq:ml_form}, one must also find the model parameters $\av$ that optimize $\obj$. This is accomplished by selecting one out of a small set of {\it algorithmic techniques}, such as stochastic gradient descent~\cite{bottou2010large}, coordinate descent~\cite{fercoq2013accelerated}, Markov Chain Monte Carlo (MCMC)\footnote{Strictly speaking, MCMC algorithms do not perform the optimization in Eq.~\ref{eq:ml_form} directly --- rather, they generate samples from the function $\obj$, and additional procedures are applied to these samples to find a optimizer $\av^*$.}~\cite{gilks2005markov}, and variational inference (to name just a few). The chosen algorithmic technique is applied to Eq.~\ref{eq:ml_form} to generate a set of {\it iterative-convergent} equations, which are implemented as program code by ML practitioners, and repeated until a {\it convergence or stopping criterion} is reached (or just as often, until a fixed computational budget is exceed). Iterative-convergent equations have the following general form:
\begin{align}
A(t) = F\left(A(t-1), \Delta_{\obj}(\av(t-1),\mathbf{x})\right)
\label{eq:master_general}
\end{align}
where the parentheses $(t)$ denotes iteration number. This general form produces the next iteration's model parameters $\av(t)$, from the previous iteration's $\av(t-1)$ and the data $\mathbf{x}$, using two functions: (1) an {\it update function} $\Delta_{\obj}$ (which increases the objective $\obj$) that performs computation on data $\mathbf{x}$ and previous model state $\av(t-1)$, and outputs intermediate results. These intermediate results are then combined to form $\av(t)$ by (2) an {\it aggregation function} $F$. For simplicity of notation, we will henceforth omit $\obj$ from the subscript of $\Delta$ --- with the implicit understanding that all ML programs considered in this paper bear an explicit loss function $\obj$ (as opposed to heuristics or procedures lacking such a loss function).

Let us now look at two concrete examples of Eqs.~\ref{eq:ml_form},~\ref{eq:master_general}, which will prove useful for understanding the unique properties of ML programs. In particular, we shall pay special attention to the 4 key components of any ML program: (1) data $\mathbf{x}$ and model $\av$; (2) loss function $f(\mathbf{x},\av)$; (3) structure-inducing function $r(\av)$; (4) algorithmic techniques than can be used for the program.

\noindent
{\bf Lasso Regression:}
Lasso regression~\cite{tibshirani1996regression} is perhaps the simplest exemplar from the structured sparse regression ML algorithm family, and is used to predict a response variable $y_i$ given vector-valued features $x_i$ (i.e. regression, which uses labeled data) --- but under the assumption that only a few dimensions or features in $x_i$ are informative about $y_i$. As input, Lasso is given $N$ training pairs $\mathbf{x}$ of the form $(\xv_i, y_i)\in\RR^m\times\RR, i=1,\ldots, n$, where the features are $m$-dimensional vectors.
The goal is to find a linear function, parametrized by the weight vector $\av$, such that (1) $\av^\top\xv_i \approx y_i$, and (2) the $m$-dimensional parameters $\av$ are sparse\footnote{Sparsity has two benefits: it automatically controls the complexity of the model (i.e. if the data requires fewer parameters, then the ML algorithm will do so), and improves human interpretation by focusing the ML practitioner's attention on just a few parameters.} (most elements are zero):
\begin{align}
\label{eq:lasso}
\min_{\av} \obj_{Lasso}(\mathbf{x},\av), \quad\mbox{ where }\quad \obj_{Lasso}(\mathbf{x},\av) = \underbrace{\frac{1}{2}\sum_{i=1}^n (\av^\top\xv_i-y_i)^2}_{f(\{x_i,y_i\}_{i=1}^N ; \av)} +  \underbrace{\lambda_n\sum_{j=1}^m |a_j|}_{r(\av)},
\end{align}
or more succinctly in matrix notation:
\begin{align}
\label{eq:lasso_m}
\min_{\av} \tfrac{1}{2}\|X\av - \yv\|_2^2 + \lambda_n\|\av\|_1,
\end{align}
where $X^\top = [\xv_1, \ldots, \xv_n]\in\RR^{m\times n}$, $\yv=(y_1, \ldots, y_n)^\top \in \RR^n$, $\|\cdot\|_2$ is the Euclidean norm on $\RR^n$, $\|\cdot\|_1$ is the $\ell_1$ norm on $\RR^m$, and $\lambda_n$ is some constant that balances model fit (the $f$ term) and sparsity (the $g$ term). Many algorithmic techniques can be applied to this problem, such as stochastic proximal gradient descent or coordinate descent. We shall present the coordinate descent\footnote{More specifically, we are presenting the form known as ``block coordinate descent", which is one of many possible forms of coordinate descent.} iterative-convergent equation:
\begin{align}
\av_j(t) = \mathbb{S}(X_{\cdot j}^\top \yv -\sum_{k \neq j} X_{\cdot j}^\top X_{\cdot k} \av_k(t-1), \lambda_n),
\label{eq:lasso_cd_update}
\end{align}
where $\mathbb{S}(\av_j,\lambda) := \mbox{sign}(\av_j)\left( \left| \av_j \right| - \lambda \right)_+$ is the ``soft-thresholding operator", and we assume the data is normalized so that for all $j$, $X_{.j}^\top X_{.j} = 1$. Tying this back to the general iterative-convergent update form, we have the following explicit forms for $\Delta,F$:
\begin{align}
\label{eq:lasso_deltaF}
\Delta_{Lasso}(\av(t-1),\mathbf{x}) & =
\begin{bmatrix}
X_{\cdot 1}^\top \yv -\sum_{k \neq 1} X_{\cdot 1}^\top X_{\cdot k} \av_k(t-1) \\
\vdots \\
X_{\cdot m}^\top \yv -\sum_{k \neq m} X_{\cdot m}^\top X_{\cdot k} \av_k(t-1)
\end{bmatrix} \\
F_{Lasso}(\av(t-1), \uv ) & =
\begin{bmatrix}
\mathbb{S}(\uv_1, \lambda_n) \\
\vdots \\
\mathbb{S}(\uv_m, \lambda_n)
\end{bmatrix}, \notag
\end{align}
where $\uv_j = [\Delta_{Lasso}(\av(t-1),\mathbf{x})]_j$ is the $j$-th element of $\Delta_{Lasso}(\av(t-1),\mathbf{x})$. 

\noindent
{\bf Latent Dirichlet Allocation Topic Model:}
Latent Dirichlet Allocation (LDA)~\cite{blei2003latent} is a member of the graphical models ML algorithm family, and is also known as a ``topic model" for its ability to identify commonly-recurring topics within a large corpus of text documents. As input, LDA is given N unlabeled documents $\mathbf{x} = \{x_i\}_{i=1}^N$, where each document $x_i$ contains $N_i$ words (referred to as ``tokens" in the LDA literature) represented by $x_i = [x_{i1},\dots,x_{ij},\dots,x_{iN_i}]$. Each token $x_{ij} \in \{1,\dots,V\}$ is an integer representing one word out of a vocabulary of $V$ words --- for example, the phrase ``machine learning algorithm" might be represented as $x_i = [x_{i1},x_{i2},x_{i3}] = [25,60,13]$ (the correspondence between words and integers is arbitrary, and has no bearing on the accuracy of the LDA algorithm).

The goal is to find a set of parameters $\av = \{\{z_{ij}\}_{i=1}^N,\{\delta_i\}_{i=1}^N,\{B_k\}_{k=1}^K\}$ --- ``token topic indicators" $z_{ij} \in \{1,\dots,K\}$ for each token in each document, ``document-topic vectors" $\delta_i \in \mathrm{Simplex}(K)$ for each document, and $K$ ``word-topic vectors" (or simply, ``topics") $B_k \in \mathrm{Simplex}(V)$ --- that maximizes the following log-likelihood\footnote{A log-likelihood is the natural logarithm of a probability distribution. As a member of the graphical models ML algorithm family, LDA specifies a probability distribution, and hence has an associated log-likelihood.} equation:
\begin{align}
\label{eq:lda}
\max_{\av} \obj_{LDA}(\mathbf{x},\av), \quad \text{where} \quad \obj_{LDA}(\mathbf{x},\av) =& \underbrace{\sum_{i=1}^N \sum_{j=1}^{N_i} \left( \ln\mathbb{P}_{Cate.}(x_{ij} \mid B_{z_{ij}}) + \ln\mathbb{P}_{Cate.}(z_{ij} | \delta_i) \right)}_{f(\{x_i\}_{i=1}^N ; \av)} \\
&+ \underbrace{\sum_{i=1}^N \ln\mathbb{P}_{Dirichlet}(\delta_i \mid \alpha) + \sum_{k=1}^K \ln\mathbb{P}_{Dirichlet}(B_k \mid \beta)}_{r(\av)}, \notag
\end{align}
where $\mathbb{P}_{Cate.}(u \mid v) \propto \prod_\ell v_\ell^{u_\ell}$ is the Categorical (a.k.a discrete) probability distribution, $\mathbb{P}_{Dirichlet}(v \mid \alpha) \propto \prod_\ell v_\ell^{\alpha - 1}$ is the Dirichlet probability distribution, and $\alpha,\beta$ are constants that balance model fit (the $f$ term) with the practitioner's prior domain knowledge about the document-topic vectors $\delta_i$ and the topics $B_k$ (the $r$ term). Similar to Lasso, many algorithmic techniques such as Gibbs sampling and variational inference (to name just two) can be used on the LDA model; we shall consider the Collapsed Gibbs sampling equations\footnote{Note that Collapsed Gibbs sampling re-represents $\delta_i,B_k$ as integer-valued vectors instead of simplex vectors. Details can be found in~\cite{yao2009efficient}.}:
\begin{align}
\label{eq:lda_update}
\forall (i,j), \quad & B_{k_{old},w_{ij}}(t-1) \mathrel{-}= 1, \\
& B_{k_{new},w_{ij}}(t-1) \mathrel{+}= 1, \notag\\
& \delta_{i,k_{old}}(t-1) \mathrel{-}= 1, \notag\\
& \delta_{i,k_{new}}(t-1) \mathrel{+}= 1, \notag\\
\text{where} \quad & k_{old} = z_{ij}(t-1) \notag\\
& k_{new} = z_{ij}(t) \sim \mathbb{P}\left(z_{ij} \mid x_{ij}, \delta_{i}(t-1), B(t-1) \right), \notag
\end{align}
where $\mathrel{+}=,\mathrel{-}=$ are the self-increment and self-decrement operators (i.e. $\delta,B,z$ are being modified in-place), $\sim \mathbb{P}()$ means ``to sample from distribution $\mathbb{P}$", and $\mathbb{P}\left(z_{ij} \mid x_{ij}, \delta_{i}(t-1), B(t-1) \right)$ is the conditional probability\footnote{There are a number of efficient ways to compute this probability. In the interest of keeping this article focused, we refer the reader to~\cite{yao2009efficient} for an appropriate introduction.} of $z_{ij}$ given the current values of $\delta_{i}(t-1),B(t-1)$. The update $\Delta_{LDA}(\av(t-1),\mathbf{x})$ proceeds in two stages: (1) execute Eq.~\ref{eq:lda_update} over all document tokens $x_{ij}$; (2) output $\av(t) = \{\{z_{ij}(t-1)\}_{i=1}^N,\{\delta_{i}(t-1)\}_{i=1}^N,\{B_{k}(t-1)\}_{k=1}^K\}$. The aggregation $F_{LDA}(\av(t-1), \dots)$ turns out to simply be the identity function.

\subsection{Unique Properties of ML Programs}
\label{sec:ml_props}

To speed up the execution of large-scale ML programs over a distributed cluster, we wish to understand their properties, which an eye towards how they can inform the design of distributed ML systems. It is helpful to first understand what an ML program is {\it not}: let us consider a traditional, non-ML program, such as sorting on MapReduce. This algorithm begins by distributing the elements to be sorted, $x_1,\dots,x_N$, randomly across a pool of $M$ mappers. The mappers hash each element $x_i$ into a key-value pair $(h(x_i),x_i)$, where $h$ is an ``order-preserving" hash function that satisfies $h(x)>h(y)$ if $x>y$. Next, for every unique key $a$, the MapReduce system sends all key-value pairs $(a,x)$ to a reducer labeled ``$a$". Each reducer then runs a sequential sorting algorithm on its received values $x$, and finally, the reducers take turns (in ascending key order) to output their sorted values.

The first thing to note about MapReduce sort, is that it is {\it single-pass} and non-iterative --- only a single Map and a single Reduce step are required. This stands in contrast to ML programs, which are iterative-convergent and repeat Eq.~\ref{eq:master_general} many times. More importantly, MapReduce sort is {\it operation-centric} and {\it deterministic}, and does not tolerate errors in individual operations: for example, if some Mapper were to output a mis-hashed pair $(a,x)$ where $a\ne h(x)$ (for the sake of argument, let us say this is due to improper recovery from a power failure), then the final output will be mis-sorted because $x$ will be output in the wrong position. It is for this reason that Hadoop and Spark (which are systems that support MapReduce) provide strong {\it operational correctness} guarantees via robust fault-tolerant systems. These fault-tolerant systems certainly require additional engineering effort, and impose additional running time overheads in the form of hard-disk-based checkpoints and lineage trees~\cite{mapreduce03,zaharia2012resilient} --- yet they are necessary for operation-centric programs, which may fail to execute correctly in their absence.

This leads us to the first property of ML programs: {\bf error tolerance}. Unlike the MapReduce sort example, ML programs are usually robust against minor errors in intermediate calculations. In Eq.~\ref{eq:master_general}, even if a limited number of updates $\Delta_\obj$ are incorrectly computed or transmitted, the ML program is still mathematically guaranteed to converge to an optimal set of model parameters $\av^*$ --- that is to say, the ML algorithm terminates with a correct output (even though it might take more iterations to do so)~\cite{ho2013more,dai2015high}. An good example is stochastic gradient descent (SGD), a frequently-used algorithmic workhorse for many ML programs, ranging from deep learning to matrix factorization and logistic regression~\cite{zhang2004solving,gemulla2011large,dean2012large}. When executing an ML program that uses SGD, even if a small random vector $\epsilon$ is added to the model after every iteration, i.e. $\av(t) = \av(t) + \epsilon$, convergence is still assured --- intuitively, this is because SGD always computes the correct direction of the optimum $\av^*$ for the update $\Delta_\obj$, moving $\av(t)$ around simply results in the direction being re-computed to suit~\cite{ho2013more,dai2015high}. This property has important implications for distributed system design, as the system no longer needs to guarantee perfect execution, inter-machine communication, or recovery from failure (which requires substantial engineering and running time overheads) --- it is often cheaper to do these approximately, especially when resources are constrained or limited (e.g. limited inter-machine network bandwidth)~\cite{ho2013more,dai2015high}.

In spite of error tolerance, ML programs can in fact be {\it harder} to execute than operation-centric programs, because of {\bf dependency structure} that is not immediately obvious from a cursory look at the objective $\obj$ or update functions $\Delta_\obj,F$. It is certainly the case that dependency structures occur in operation-centric programs: in MapReduce sort, the reducers must wait for the mappers to finish, otherwise the sort will be incorrect. In order to see what makes ML dependency structures unique, let us consider the Lasso regression example in Eq.~\ref{eq:lasso}: at first glance, the $\Delta_{Lasso}$ update equations~\ref{eq:lasso_deltaF} may look like they can be executed in parallel, but this is only partially true. A more careful inspection reveals that, for the $j$-th model parameter $\av_j$, its update depends on $\sum_{k \neq j} X_{\cdot j}^\top X_{\cdot k} \av_k(t-1)$ --- in other words, potentially every other parameter $\av_k$ is a possible dependency, and therefore the order in which the model parameters $\av$ are updated has an impact on the ML program's progress or even correctness~\cite{lee2014model}. Even more, there is an additional nuance not present in operation-centric programs: the Lasso parameter dependencies are not binary (i.e. only on or off), but can be soft-valued and influenced by both the ML program state and input data: notice that if $X_{\cdot j}^\top X_{\cdot k} = 0$ (meaning that data column $j$ is uncorrelated with column $k$), then $\av_j$ and $\av_k$ have zero dependency on each other, and can be updated safely in parallel~\cite{lee2014model}. Similarly, even if $X_{\cdot j}^\top X_{\cdot k} > 0$, as long as $\av_k = 0$, then $\av_j$ does not depend on $\av_k$. Such dependency structures are not limited to one ML program; careful inspection of the LDA topic model update equations~\ref{eq:lda_update} reveals the Gibbs sampler update for $x_{ij}$ (word token $j$ in document $i$) depends on (1) all other word tokens in document $i$, and (2) all other word tokens $b$ in other documents $a$ that represent the exact same word, i.e. $x_{ij} = x_{ab}$~\cite{zheng2015model}. If these ML program dependency structures are not respected, the result is either sub-ideal scaling with additional machines (e.g. $<2\times$ speedup with $4\times$ as many machines)~\cite{zheng2015model} or even outright program failure that overwhelms the intrinsic error tolerance of ML programs~\cite{lee2014model}.

A third property of ML programs is {\bf non-uniform convergence}, the observation that not all model parameters $\av_j$ will converge to their optimal values $\av_j^*$ in the same number of iterations --- a property that is absent from single-pass algorithms like MapReduce sort. In the Lasso example Eq.~\ref{eq:lasso}, the $r(A)$ term encourages model parameters $\av_j$ to be exactly zero, and it has been empirically observed that once a parameter reaches zero during algorithm execution, it is unlikely to revert to a non-zero value~\cite{lee2014model} --- to put it another way, parameters that reach zero are (with high, though not $100\%$, probability) already converged. This suggests that computation may be better prioritized towards parameters that are still non-zero, by executing $\Delta_{Lasso}$ more frequently on them --- and such a strategy indeed reduces the time taken by the ML program to finish~\cite{lee2014model}. Similar non-uniform convergence has been observed and exploited in PageRank, another iterative-convergent algorithm~\cite{graphlab10}.

Finally, it is worth noting that a subset of ML programs exhibit {\bf compact updates}, in that the updates $\Delta_{Lasso}$ are, upon careful inspection, {\it significantly smaller than} the size of the matrix parameters, $|\av|$. In both Lasso (Eq.~\ref{eq:lasso}) and LDA topic models~\cite{blei2003latent}, the updates $\Delta_{Lasso}$ generally touch just a small number of model parameters, due to sparse structure in the data. Another salient example is that of ``matrix-parametrized" models, where $\av$ is a matrix (such as in deep learning~\cite{hinton2006reducing}), yet individual updates $\Delta_{Lasso}$ can be decomposed into a few small vectors (a so-called ``low-rank" update). Such compactness can dramatically reduce storage, computation, and communication costs if the distributed ML system is designed with it in mind, resulting in order-of-magnitude speedups~\cite{xie2015distributed,zhang2015poseidon}.

\subsection{On Data and Model Parallelism}

\begin{figure}[t]
\centering
\includegraphics[width=0.8\linewidth]{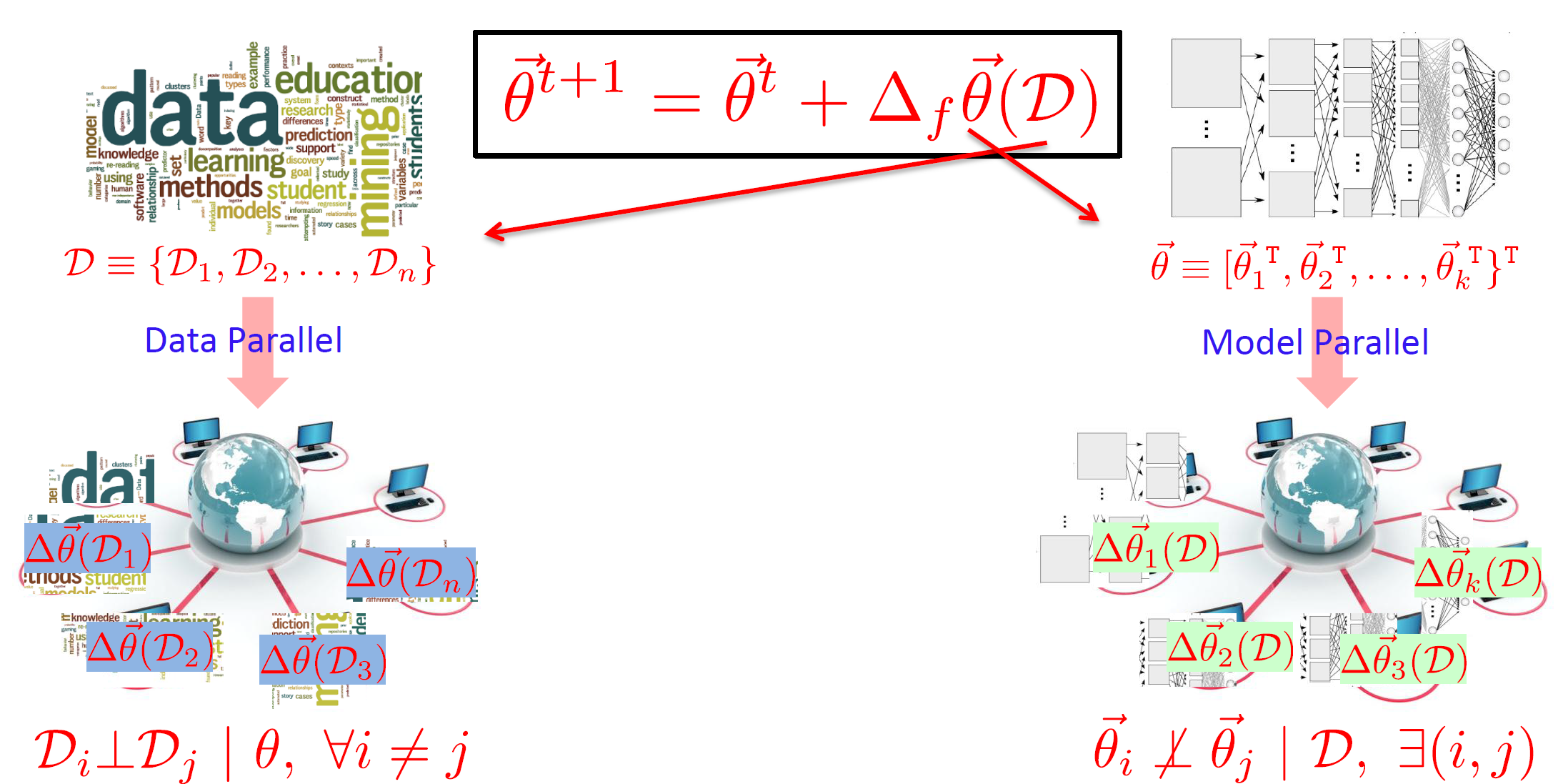}
\caption{\small The difference between data and model parallelism: data samples are always conditionally independent given the model, but there are some model parameters that are not independent of each other.}
\label{fig:data_model_parallel_difference}
\end{figure}

For ML applications involving terabytes of data, using complex ML programs with up to trillions of model parameters, execution on a single desktop or laptop often takes days or weeks~\cite{krizhevsky2012imagenet}; this computational bottleneck has spurred the development of many distributed systems for parallel execution of ML programs over a cluster~\cite{gonzalez2012powergraph,zaharia2012resilient,li2014scaling,xing2015petuum}. ML programs are parallelized by subdividing the updates $\Delta_\obj$ over either the data $\mathbf{x}$ or the model $\av$ --- referred to respectively as {\it data parallelism} and {\it model parallelism}.

It is crucial to note that the two types of parallelism are complementary and asymmetric --- complementary, in that simultaneous data and model parallelism is possible (and even necessary, in some cases), and asymmetric, in that data parallelism can be applied generically to any ML program with an {\it independent and identically distributed} (i.i.d.) assumption over the data samples $x_1,\dots,x_N$; such i.i.d. ML programs (from deep learning, to logistic regression, to topic modeling and many others) make up the bulk of practical ML usage, and are easily recognized by a summation over data indices $i$ in the objective $\obj$ (for example, Lasso Eq.~\ref{eq:lasso}). Consequently, when a workhorse algorithmic technique (e.g. stochastic gradient descent) is applied to $\obj$, the derived update equations $\Delta_\obj$ will also have a summation\footnote{For Lasso coordinate descent $\Delta_{Lasso}$ (Eq.~\ref{eq:lasso_cd_update}), the summation over $i$ is in the inner product $X_{\cdot j}^\top X_{\cdot k}=\sum_{i=1}^N X_{ij}X_{ik}$.} over $i$, which can be easily parallelized over multiple machines, particularly when the number of data samples $N$ is in the millions or billions. In contrast, model parallelism requires special care, because model parameters $\av_j$ do not always enjoy this convenient i.i.d assumption (Figure~\ref{fig:data_model_parallel_difference}) --- therefore, {\it which} parameters $\av_j$ are updated in parallel, as well as the {\it order} in which the updates $\Delta_\obj$ happen, can lead to a variety of outcomes: from near-ideal $P$-fold speedup with $P$ machines, to no additional speedups with additional machines, or even to complete program failure. The dependency structures discussed for Lasso (Section~\ref{sec:ml_props}) are a good example of the non-i.i.d. nature of model parameters. Let us now discuss the general mathematical forms of data and model parallelism, respectively.

\noindent
{\bf Data Parallelism: }
In data parallel ML execution, the data $\mathbf{x}=\{x_1,\dots,x_N\}$ is partitioned and assigned to parallel computational workers or machines (indexed by $p=1,\dots,P$); we shall denote the $p$-th data partition by $\mathbf{x}_p$. If the update function $\Delta_\obj$ has an outermost summation over data samples $i$ (as seen in ML programs with the commonplace i.i.d. assumption on data), we can split $\Delta_\obj$ over data subsets and obtain a data parallel update equation, in which $\Delta_\obj(\av(t-1),\mathbf{x}_p)$ is executed on the $p$-th parallel worker:
\begin{equation}
\textstyle \av(t) = F(\av(t-1), \sum_{p=1}^P \Delta_\obj(\av(t-1),\mathbf{x}_p) ).
\label{eq:data_parallel}
\end{equation}
It is worth noting that the summation $\sum_{p=1}^P$ is the basis for a host of established techniques for speeding up data-parallel execution, such as minibatches and bounded-asynchronous execution~\cite{ho2013more,dai2015high}. As a concrete example, we can write the Lasso block coordinate descent equations~\ref{eq:lasso_deltaF} in a data parallel form, by applying a bit of algebra:
\begin{align}
\label{eq:data_parallel_lasso}
\Delta_{Lasso}(\av(t-1),\mathbf{x}_p) & =
\begin{bmatrix}
\sum_{i\in\mathbf{x}_p} \left( X_{i1} \yv_i -\sum_{k \neq 1} X_{i1} X_{ik} \av_k(t-1) \right) \\
\vdots \\
\sum_{i\in\mathbf{x}_p} \left( X_{im} \yv_i -\sum_{k \neq m} X_{im} X_{ik} \av_k(t-1) \right)
\end{bmatrix} \\
F_{Lasso}(\av(t-1), \uv ) & =
\begin{bmatrix}
\mathbb{S}\left(\left[\sum_{p=1}^P \Delta_{Lasso}(\av(t-1),\mathbf{x}_p)\right]_1, \lambda_n\right) \\
\vdots \\
\mathbb{S}\left(\left[\sum_{p=1}^P \Delta_{Lasso}(\av(t-1),\mathbf{x}_p)\right]_m, \lambda_n\right)
\end{bmatrix}, \notag
\end{align}
where $\sum_{i\in\mathbf{x}_p}$ means (with a bit of notation abuse) to sum over all data indices $i$ included in $\mathbf{x}_p$.

\noindent
{\bf Model Parallelism:} In model parallel ML execution, the model $A$ is partitioned and assigned to workers/machines $p=1,\dots,P$, and updated therein by running parallel update functions $\Delta_\obj$. Unlike data-parallelism, each update function $\Delta_\obj$ also takes a {\it scheduling} or selection function $S_{p,(t-1)}$, which restricts $\Delta_\obj$ to operate on a subset of the model parameters $\av$ (one basic use is to prevent different workers from trying to update the same parameters):
\begin{equation}
\av(t) = F\left(\av(t-1), \{ \Delta_\obj(\av(t-1), S_{p,(t-1)}(\av(t-1))) \}_{p=1}^P \right),
\label{eq:model_parallel}
\end{equation}
where we have omitted the data $\mathbf{x}$ since it is not being partitioned over. $S_{p,(t-1)}$ outputs a set of indices $\{j_1,j_2,\dots,\}$, so that $\Delta_\obj$ only performs updates on $\av_{j_1},\av_{j_2},\dots$ --- we refer to such selection of model parameters as {\it scheduling}. The model parameters $A_j$ are not, in general, independent of each other, and it has been established that model parallel algorithms are effective only when each iteration of parallel updates is restricted to a subset of {\it mutually independent} (or weakly-correlated) parameters~\cite{lee2014model,shotgun,feature_cluster,graphlab12}, which can be performed by $S_{p,(t-1)}$.

\begin{figure}[t]
\centering
\includegraphics[width=0.75\linewidth]{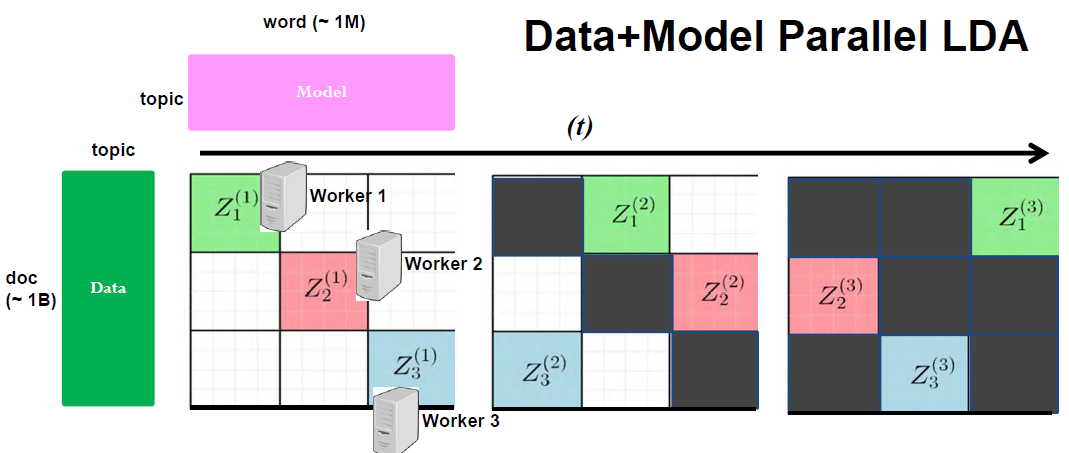}
\caption{\small High-level illustration of simultaneous data and model parallelism in LDA topic modeling. In this example, the 3 parallel workers operate on data/model blocks $Z_1^{(1)},Z_2^{(1)},Z_3^{(1)}$ during iteration 1, then move on to blocks $Z_1^{(2)},Z_2^{(2)},Z_3^{(2)}$ during iteration 2, and so forth.}
\label{fig:data_model_parallel_lda}
\end{figure}

The Lasso block coordinate descent updates (Eq.~\ref{eq:lasso_deltaF}) can be easily written in a simple model parallel form. Here, $S_{p,(t-1)}$ chooses the same fixed set of parameters for worker $p$ on every iteration, which we refer to by $j_{p1},\dots,j_{pm_p}$:
\begin{align}
\label{eq:model_parallel_lasso}
\Delta_{Lasso}(\av(t-1), S_{p,(t-1)}(\av(t-1))) & =
\begin{bmatrix}
X_{\cdot j_{p1}}^\top \yv -\sum_{k \neq j_{p1}} X_{\cdot j_{p1}}^\top X_{\cdot k} \av_k(t-1) \\
\vdots \\
X_{\cdot j_{pm_p}}^\top \yv -\sum_{k \neq j_{pm_p}} X_{\cdot j_{pm_p}}^\top X_{\cdot k} \av_k(t-1)
\end{bmatrix} \\
F_{Lasso}(\av(t-1), \dots ) & =
\begin{bmatrix}
\mathbb{S}\left(\left[\Delta_{Lasso}(\av(t-1), S_{1,(t-1)}(\av(t-1))) \right]_1, \lambda_n\right) \\
\vdots \\
\mathbb{S}\left(\left[\Delta_{Lasso}(\av(t-1), S_{1,(t-1)}(\av(t-1))) \right]_{m_1}, \lambda_n\right) \\
\vdots \\
\vdots \\
\mathbb{S}\left(\left[\Delta_{Lasso}(\av(t-1), S_{P,(t-1)}(\av(t-1))) \right]_1, \lambda_n\right) \\
\vdots \\
\mathbb{S}\left(\left[\Delta_{Lasso}(\av(t-1), S_{P,(t-1)}(\av(t-1))) \right]_{m_P}, \lambda_n\right) \notag
\end{bmatrix}.
\end{align}
On a closing note, simultaneous data and model parallelism is also possible, by partitioning the space of data samples and model parameters $(x_i,\av_j)$ into disjoint blocks. The LDA topic model Gibbs sampling equations (Eq.~\ref{eq:lda_update}) can be partitioned in such a block-wise manner (Figure~\ref{fig:data_model_parallel_lda}), in order to achieve near-perfect speedup with $P$ machines~\cite{zheng2015model}.

%% file: principles.tex

\section{Principles of ML System Design}

The unique properties of ML programs, when coupled with the complementary strategies of data and model parallelism, interact to produce a complex space of design considerations that goes beyond the ideal mathematical view suggested by the general iterative-convergent update equation Eq.~\ref{eq:master_general}. In this ideal view, one hopes that the $\Delta,F$ functions simply need to be implemented equation-by-equation (e.g. following the Lasso regression data and model parallel equations earlier), and then executed by a general purpose distributed system --- for example, if we chose a MapReduce abstraction, one could write $\Delta$ as Map and $F$ as Reduce, and then use a system such as Hadoop or Spark to execute them. The reality, however, is that the highest-performing ML implementations are {\it not} built in such a naive manner, and furthermore, they tend to be found in ML-specialized systems rather than on general-purpose MapReduce systems~\cite{hogwild,li2014scaling,xing2015petuum,yuan2015lightlda}. The reason is that high-performance ML goes far beyond an idealized MapReduce-like view, and involves numerous considerations that are not immediately obvious from the mathematical equations: considerations such as what data batch size to use for data parallelism, how to partition the model for model parallelism, when to synchronize model views between workers, step size selection for gradient based algorithms, and even the order in which to perform $\Delta$ updates.

The space of ML performance considerations can be intimidating to even veteran practitioners, and it is our view that a {\it systems interface} for parallel ML is needed, both to (a) facilitate the organized, scientific study of ML considerations, and also to (b) organize these considerations into a series of high-level principles for developing new distributed ML systems. As a first step towards organizing these principles, we shall divide them according to 4 high-level questions: if an ML program's equations (Eq.~\ref{eq:master_general}) tell the system ``what to compute", then the system must consider: (1) How to distribute the computation? (2) How to bridge computation with inter-machine communication? (3) How to communicate between machines? (4) What to communicate? By systematically addressing the ML considerations that fall under each question, we show that it is possible to build sub-systems whose benefits complement and accrue with each other, and which can be assembled into a full distributed ML system that enjoys orders-of-magnitude speedups in ML program execution time.

\subsection{How to Distribute: Scheduling and Balancing workloads}
\label{sec:how_to_distribute}

In order to parallelize an ML program, we must first determine how best to {\it partition} it into multiple tasks --- that is to say, we must partition the monolithic $\Delta$ in Eq.~\ref{eq:master_general} into a set of parallel tasks, following the data parallel form (Eq.~\ref{eq:data_parallel}) or the model parallel form (Eq.~\ref{eq:model_parallel}) --- or even a more sophisticated hybrid of both forms. Then, we must schedule and balance those tasks for execution on a limited pool of $P$ workers or machines: that is to say, we decide (i) which tasks go together in parallel (and just as importantly, which tasks should {\it not} be executed in parallel), (ii) the order in which tasks will be executed, while simultaneously ensuring (iii) each machine's share of the workload is well-balanced.

These three decisions have been carefully studied in the context of operation-centric programs (such as the MapReduce sort example), giving rise (for example) to the scheduler system used in Hadoop and Spark~\cite{zaharia2012resilient}. Such operation-centric scheduler systems may come up with a different {\it execution plan} --- the combination of decisions (i)-(iii) --- depending on the cluster configuration, existing workload, or even machine failure; yet, crucially, they ensure that the outcome of the operation-centric program is perfectly consistent and reproducible every time. However, for ML iterative-convergent programs, the goal is not perfectly reproducible execution, but rather convergence of the model parameters $\av$ to an optimum of the objective function $\obj$ (that is to say, $\av$ approaches to within some small distance $\epsilon$ of an optimum $\av^*$). Accordingly, we would like to develop a scheduling strategy whose execution plans allow ML programs to {\it provably} terminate with the same quality of convergence every time --- we shall refer to this as ``correct execution" for ML programs. Such a strategy can then be implemented as a scheduling system, which creates ML program execution plans that are distinct from operation-centric ones.

\noindent
{\bf Dependency Structures in ML Programs:}
In order to generate a correct execution plan for ML programs, it is necessary to understand how ML programs have internal dependencies, and how breaking or {\it violating} these dependencies through naive parallelization will slow down convergence. Unlike operation-centric programs such as sorting, ML programs are error-tolerant, and can automatically recover from a limited number of dependency violations --- but too many violations will increase the number of iterations required for convergence, and cause the parallel ML program to experience {\it sub-optimal}, less-than-$P$-fold speedup with $P$ machines.

Let us understand these dependencies through the Lasso and LDA topic model example programs. In the model parallel version of Lasso (Eq.~\ref{eq:model_parallel_lasso}), each parallel worker $p\in\{1,\dots,P\}$ performs one or more $\Delta_{Lasso}$ calculations of the form $X_{\cdot j}^\top \yv -\sum_{k \neq j} X_{\cdot j}^\top X_{\cdot k} \av_k(t-1)$, which will then be used to update $\av_j$. Observe that this calculation depends on {\it all other} parameters $\av_k$, $k\ne j$ through the term $X_{\cdot j}^\top X_{\cdot k} \av_k(t-1)$, with the magnitude of the dependency being proportional to (1) the correlation between the $j$-th and $k$-th data dimensions, $X_{\cdot j}^\top X_{\cdot k}$; (2) the current value of parameter $\av_k(t-1)$. In the worst case, both the correlation $X_{\cdot j}^\top X_{\cdot k}$ and $\av_k(t-1)$ could be large, and therefore updating $\av_j,\av_k$ sequentially (that is to say, over two different iterations $t$, $t+1$) will lead to a different result from updating them in parallel (i.e. at the same time in iteration $t$). \cite{shotgun} noted that, if the correlation is large, then the parallel update will take more iterations to converge than the sequential update. It intuitively follows that we should not ``waste" computation trying to update highly correlated parameters in parallel --- rather we should seek to schedule {\it uncorrelated} groups of parameters for parallel updates, while performing updates for correlated parameters sequentially~\cite{lee2014model}.

For LDA topic modeling, let us recall the $\Delta_{LDA}$ updates (Eq.~\ref{eq:lda_update}): for every word token $w_{ij}$ (in position $j$ in document $i$), the LDA Gibbs sampler updates 4 elements of the model parameters $B,\delta$ (which are part of $\av$): $B_{k_{old},w_{ij}}(t-1) \mathrel{-}= 1$, $B_{k_{new},w_{ij}}(t-1) \mathrel{+}= 1$, $\delta_{i,k_{old}}(t-1) \mathrel{-}= 1$, $\delta_{i,k_{new}}(t-1) \mathrel{+}= 1$, where $k_{old} = z_{ij}(t-1)$ and $k_{new} = z_{ij}(t) \sim \mathbb{P}\left(z_{ij} \mid x_{ij}, \delta_{i}(t-1), B(t-1) \right)$. These equations give rise to many dependencies between different word tokens $w_{ij}$ and $w_{uv}$; one obvious dependency occurs when $w_{ij}=w_{uv}$, leading to a chance that they will update the same elements of $B$ (which happens when $k_{old}$ or $k_{new}$ are the same for both tokens). Furthermore, there are more complex dependencies inside the conditional probability $\mathbb{P}\left(z_{ij} \mid x_{ij}, \delta_{i}(t-1), B(t-1) \right)$; in the interest of keeping this article at a suitably high level, we will summarize by noting that elements in the columns of $B$, i.e. $B_{\cdot,v}$, are mutually dependent, while elements in the rows of $\delta$, i.e. $\delta_{i,\cdot}$, are also mutually dependent. Due to these intricate dependencies, high-performance parallelism of LDA topic modeling requires a simultaneous data-and-model parallel strategy (Figure~\ref{fig:data_model_parallel_lda}), where word tokens $w_{ij}$ must be carefully grouped by both their value $v=w_{ij}$ and their document $i$, which avoids violating the column/row dependencies in $B,\delta$~\cite{zheng2015model}.

\noindent
{\bf Scheduling in ML Programs:}
In light of these dependencies, how can we schedule the updates $\Delta$ in a manner that avoids violating as many dependency structures as possible (noting that we do not have to avoid {\it all} dependencies thanks to ML error tolerance) --- yet, at the same time, does not leave any of the $P$ worker machines idle due to lack of tasks or poor load balance? These two considerations have distinct yet complementary effects on ML program execution time: avoiding dependency violations prevents the {\it progress per iteration} of the ML program from degrading compared to sequential execution (i.e. the program will not need more iterations to converge), while keeping worker machines fully occupied with useful computation ensures that the {\it iteration throughput} (iterations executed per second) from $P$ machines is as close to $P$ times that of a single machine. In short, near-perfect $P$-fold ML speedup results from combining near-ideal progress per iteration (equal to sequential execution) with near-ideal iteration throughput ($P$ times sequential execution) --- thus, we would like to have an {\it ideal} ML scheduling strategy that attains these two goals.

\begin{figure}[t]
\begin{center}
\includegraphics[width=0.6\textwidth]{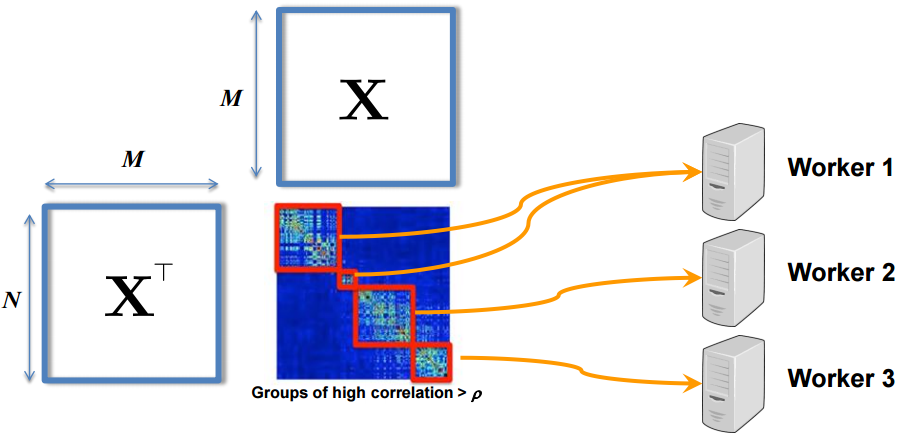}
\caption{Illustration of ideal Lasso scheduling, in which parameter pairs $(j,k)$ are grouped into subsets (red blocks) with low correlation between parameters in different subsets. Multiple subsets can be updated in parallel by multiple worker machines; this avoids violating dependency structures because workers update the parameters in each subset {\it sequentially}.}
\label{fig:lasso}
\end{center}
\end{figure}

To explain how ideal scheduling can be realized, we return to our running Lasso and LDA examples. In Lasso, the degree to which two parameters $\av_j,\av_k$ are interdependent is influenced by the data correlation $X_{\cdot j}^\top X_{\cdot k}$ between the $j$-th and $k$-th feature dimensions --- we refer to this and other similar operations as a {\it dependency check}. If $X_{\cdot j}^\top X_{\cdot k} < \kappa$ for a small threshold $\kappa$, then $\av_j,\av_k$ will have little influence on each other. Hence, the ideal scheduling strategy is to find all pairs $(j,k)$ such that $X_{\cdot j}^\top X_{\cdot k} < \kappa$, and then partition the parameter indices $j\in\{1,\dots,m\}$ into {\it independent subsets} $\mathbf{A}_1,\mathbf{A}_2,\dots$ --- where two subsets $\mathbf{A}_a,\mathbf{A}_b$ are said to be independent if for any $j\in\mathbf{A}_a$ and any $k\in\mathbf{A}_b$, we have $X_{\cdot j}^\top X_{\cdot k} < \kappa$. These subsets $\mathbf{A}$ can then be safely assigned to parallel worker machines (Fig.(\ref{fig:lasso})), and each machine will update the parameters $j\in\mathbf{A}$ sequentially (thus preventing dependency violations)~\cite{lee2014model}.

As for LDA, careful inspection reveals that the update equations $\Delta_{LDA}$ for word token $w_{ij}$ (Eq.~\ref{eq:lda_update}) may (1) touch any element of column $B_{\cdot,w_{ij}}$, and (2) touch any element of row $\delta_{i,\cdot}$. In order to prevent parallel worker machines from operating on the same columns/rows of $B,\delta$, we must partition the space of words $\{1,\dots,V\}$ (corresponding to columns of $B$) into $P$ subsets $\mathbf{V}_1,\dots,\mathbf{V}_P$, as well as partition the space of documents $\{1,\dots,N\}$ (corresponding to rows of $\delta$) into $P$ subsets $\mathbf{D}_1,\dots,\mathbf{D}_P$. We may now perform ideal data-and-model parallelization as follows: first, we assign document subset $\mathbf{D}_p$ to machine $p$ out of $P$. Then, each machine $p$ will only Gibbs sample word tokens $w_{ij}$ such that $i\in\mathbf{D}_p$ and $w_{ij}\in\mathbf{V}_p$. Once all machines have finished, they {\it rotate} word subsets $\mathbf{V}_p$ amongst each other, so that machine $p$ will now Gibbs sample $w_{ij}$ such that $i\in\mathbf{D}_p$ and $w_{ij}\in\mathbf{V}_{p+1}$ (or for machine $P$, $w_{ij}\in\mathbf{V}_1$). This process continues until $P$ rotations have completed, at which point the iteration is complete (every word token has been sampled)~\cite{zheng2015model}. Figure~\ref{fig:data_model_parallel_lda} illustrates this process.

In practice, ideal schedules like the ones above may not be practical to use. For instance, in Lasso, computing $X_{\cdot j}^\top X_{\cdot k}$ for all $\mathcal{O}(m^2)$ pairs $(j,k)$ is intractable for high dimensional problems with large $m$ (millions to billions). We will return to this issue shortly, when we introduce Structure Aware Parallelization (SAP), a provably near-ideal scheduling strategy that can be computed quickly.

\noindent
{\bf Compute Prioritization in ML Programs:}
Because ML programs exhibit non-uniform parameter convergence, an ML scheduler has an opportunity to prioritize slower-to-converge parameters $\av_j$, thus improving the progress per iteration of the ML algorithm (i.e. requires fewer iterations to converge). For example, in Lasso, it has been empirically observed that the sparsity-inducing $\ell_1$ norm (Eq.~\ref{eq:lasso_m}) causes most parameters $\av_j$ to (1) become exactly zero after a few iterations, after which (2) they are unlikely to become non-zero again. The remaining parameters, which are typically a small minority, take much longer to converge (such as 10 times more iterations)~\cite{lee2014model}.

A general yet effective prioritization strategy is to select parameters $\av_j$ with probability proportional to their squared rate of change, $(\av_j(t-1)-\av_j(t-2))^2 + \epsilon$ --- where $\epsilon$ is a small constant that ensures stationary parameters still have a small chance to be selected. Depending on the ratio of fast- to slow-converging parameters, this prioritization strategy can an order-of-magnitude reduction in the number of iterations required to converge by Lasso regression~\cite{lee2014model}. Similar strategies have been applied to PageRank, another iterative-convergent algorithm~\cite{graphlab10}.

\noindent
{\bf Balancing Workloads in ML Programs:}
When executing ML programs over a distributed cluster, they may have to stop in order to exchange parameter updates, i.e. {\it synchronize} --- for example, at the end of Map or Reduce phases in Hadoop and Spark. In order to reduce the time spent waiting, it is desirable to {\it load-balance} the work on each machine, so that they proceed at close to the same rate. This is especially important for ML programs, which may exhibit skewed data distributions: for example, in LDA topic models, the word tokens $w_{ij}$ are distributed in a power-law fashion, where a few words occur across many documents, while most other words appear rarely. A typical ML load-balancing strategy might apply the classic bin packing algorithm from computer science (where each worker machine is one of the ``bins" to be packed), or any other strategy that works for operation-centric distributed systems such as Hadoop and Spark.

However, a second, less-appreciated challenge is that machine performance may fluctuate in real-world clusters, due to subtle reasons such as changing datacenter temperature, machine failures, background jobs, or other users. Thus, load balancing strategies that are predetermined at the start of an iteration will often suffer from {\it stragglers}, machines that randomly become slower than the rest of the cluster, and which all other machines must wait for when performing parameter synchronization at the end of an iteration~\cite{ho2013more,dai2015high,chilimbi2014project}. An elegant solution to this problem is to apply {\it slow-worker agnosticism}~\cite{kumar2014fugue}, where the system takes direct advantage of the iterative-convergent nature of ML algorithms, and allows the faster workers to repeat their updates $\Delta$ whilst waiting for the stragglers to catch up. This not only solves the straggler problem, but can even correct for imperfectly-balanced workloads. We note that another solution to the straggler problem is to use bounded-asynchronous execution (as opposed to synchronous MapReduce-style execution) --- we shall discuss this in more detail in Section~\ref{sec:bridging}.

\noindent
{\bf Structure Aware Parallelization:}
\begin{figure}[t]
\begin{center}
\includegraphics[width=0.8\textwidth]{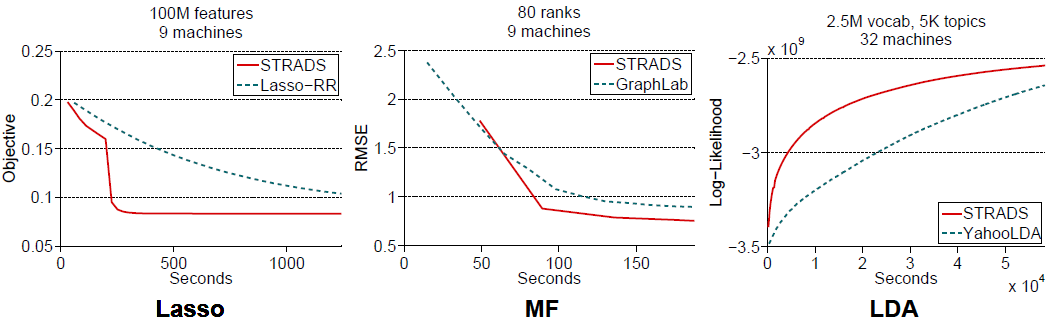}
\caption{(Adapted from~\cite{lee2014model}) Objective function $\obj$ progress versus time plots for three ML programs --- Lasso Regression, Matrix Factorization, Latent Dirichlet Allocation topic modeling --- executed under Strads, a system that realizes the Structure Aware Parallelization (SAP) abstraction. By using SAP to improve progress per iteration of ML algorithms, Strads achieves faster time to convergence (steeper curves) than other general- and special-purpose implementations --- Lasso-RR (a.k.a Shotgun algorithm), GraphLab and YahooLDA.}
\label{fig:sap_strads}
\end{center}
\end{figure}
Scheduling, prioritization and load-balancing are complementary yet intertwined --- the choice of parameters $\av_j$ to prioritize will influence which dependency checks the scheduler needs to perform, and in turn, the ``independent subsets" produced by the scheduler can make the load-balancing problem more or less difficult. These three functionalities can be combined into a single programmable abstraction, to be implemented as part of a distributed system for ML. We call this abstraction Structure Aware Parallelization (SAP),
in which ML programmers can specify how to (1) prioritize parameters to speed up convergence; (2) perform dependency checks on the prioritized parameters, and schedule them into {\it independent subsets}; (3) load-balance the independent subsets across the worker machines.
SAP exposes a simple, MapReduce-like programming interface, where ML programmers implement three functions: (1) \texttt{schedule()}, in which a small number of parameters are prioritized, and then exposed to dependency checks; (2) \texttt{push()}, which performs $\Delta_{\obj}$ in parallel on worker machines; (3) \texttt{pull()}, which performs $F$. Load balancing is automatically handled by the SAP implementation, through a combination of classic bin packing and slow-worker agnosticism.

Importantly, SAP \texttt{schedule()} does not naively perform $\mathcal{O}(m^2)$ dependency checks --- instead, a few parameters $\mathbf{A}$ are first selected via prioritization (where $|\mathbf{A}| \ll m$). The dependency checks are then performed on $\mathbf{A}$, and the resulting independent subsets are updated via \texttt{push()} and \texttt{pull()}. Thus, SAP only updates a few parameters $\av_j$ per iteration of \texttt{schedule()}, \texttt{push()}, \texttt{pull()}, rather than the full model $\av$. This strategy is provably near-ideal for a broad class of model parallel ML programs:
\begin{theorem}[adapted from~\cite{xing2015petuum}]
{\bf SAP is close to ideal execution:}
Consider objective functions of the form $\obj = f(\av) + r(\av)$, where $r(\av) = \sum_j r(A_j)$ is separable, $\av\in\RR^d$, and $f$ has $\beta$-Lipschitz continuous gradient in the following sense:
\begin{align}
\label{eq:lcg}
f(\av + \zvec) \leq f(\av) + \zvec^\top \nabla f(\av) + \tfrac{\beta}{2}\av^\top X^\top X \zvec.
\end{align}
Let $X = [\xvec_1, \ldots, \xvec_d]$ be the data samples re-represented as $d$ feature vectors. W.l.o.g., we assume that each feature vector $\xvec_i$ is normalized, i.e., $\|\xvec_i\|_2  = 1, i = 1,\ldots, d$. Therefore $ |\xvec_i^\top \xvec_j| \leq 1$ for all $i, j$.

Suppose we want to minimize $\obj$ via model parallel coordinate descent. Let $S_{ideal}()$ be an oracle (i.e. ideal) schedule that always proposes $P$ random features with zero correlation. Let $\av_{ideal}^{(t)}$ be its parameter trajectory, and let $\av_{SAP}^{(t)}$ be the parameter trajectory of SAP scheduling. Then,
\begin{align}
\EE[|\av_{ideal}^{(t)} &- \av_{SAP}^{(t)}|] \leq \frac{2dPm}{(t+1)^2\hat{P}} L^2 X^\top XC,
\end{align}
for constants $C,m,L,\hat{P}$.
\end{theorem}
This theorem says that the difference between the $S_{SAP}()$ parameter estimate $\av_{SAP}$ and the ideal oracle estimate $\av_{ideal}$ rapidly vanishes, at a fast $1/(t+1)^2 = \mathcal{O}(t^{-2})$ rate. In other words, one cannot do much better than $S_{SAP}()$ scheduling --- it is near-optimal.

SAP's slow-worker agnostic load-balancing also comes with a theoretical performance guarantee --- it not only preserves correct ML convergence, but also improves convergence per iteration over naive scheduling:
\begin{theorem}[adapted from~\cite{kumar2014fugue}]
{\bf SAP slow-worker agnosticism improves convergence progress per iteration:}
Let the current variance (intuitively, the uncertainty) in the model be $\Var(\av)$, and let $n_p>0$ be the number of updates performed by worker $p$ (including additional updates due to slow-worker agnosticism). After $n_p$ updates, $\Var(\av)$ is reduced to
\begin{align}
\Var(\av^{+n_p}) = \Var(\av) - c_{1}\eta_t n_p\Var(\av) - c_{2}\eta_t n_p\mathrm{CoVar}(\av,\nabla\obj) + c_{3}\eta_t^2 n_p + \mathcal{O}(\mathrm{cubic}),
\end{align}
where $\eta_t>0$ is a step-size parameter that approaches zero as $t\rightarrow\infty$, $c_1,c_2,c_3>0$ are problem-specific constants, $\nabla\obj$ is the stochastic gradient of the ML objective function $\obj$, $\mathrm{CoVar}(a,b)$ is the covariance between $a,b$, and $\mathcal{O}(\mathrm{cubic})$ represents 3rd-order and higher terms that shrink rapidly towards zero.
\end{theorem}
A low variance $\Var(\av)$ indicates that the ML program is close to convergence (because the parameters $\av$ have stopped changing quickly). The above theorem shows that additional updates $n_p$ do indeed lower the variance --- therefore, the convergence of the ML program is accelerated. To see why this is the case, we note that the 2nd and 3rd terms are always negative; furthermore, they are $\mathcal{O}(\eta_t)$, so they dominate the 4th positive term (which is $\mathcal{O}(\eta_t^2)$ and therefore shrinks towards zero faster) as well as the 5th positive term (which is 3rd-order and shrinks even faster than the 4th term).

Empirically, SAP systems achieve order-of-magnitude speedups over non-scheduled and non-balanced distributed ML systems. One example is the Strads system~\cite{lee2014model}, which implements SAP schedules for several algorithms, such as Lasso Regression, Matrix Factorization, and Latent Dirichlet Allocation topic modeling, and achieves superior convergence times compared to other systems (Fig. ~\ref{fig:sap_strads}).

\subsection{How to Bridge Computation and Communication:\\
Bridging Models and Bounded Asynchrony}
\label{sec:bridging}

Many parallel programs require worker machines to exchange program state between each other --- for example, MapReduce systems like Hadoop take the key-value pairs $(a,b)$ created by all Map workers, and transmit all pairs with key $a$ to the same Reduce worker. For operation-centric programs, this step must be executed perfectly without error --- recall the MapReduce sort example (Section~\ref{sec:background}), where sending keys to two different reducers results in a sorting error. This notion of operational-correctness in parallel programming is underpinned by Bulk Synchronous Parallel (BSP)~\cite{valiant1990bridging,mccoll1995bulk}, a {\it bridging model} that provides an abstract view of how parallel program computations are interleaved with inter-worker communication. Programs that follow the BSP bridging model alternate between a computation phase, and a communication phase or {\it synchronization barrier} (Figure~\ref{fig:bsp}), and the effects of each computation phase are not visible to worker machines until the next synchronization barrier has completed.

\begin{figure}[t]
\begin{center}
\includegraphics[width=0.6\textwidth]{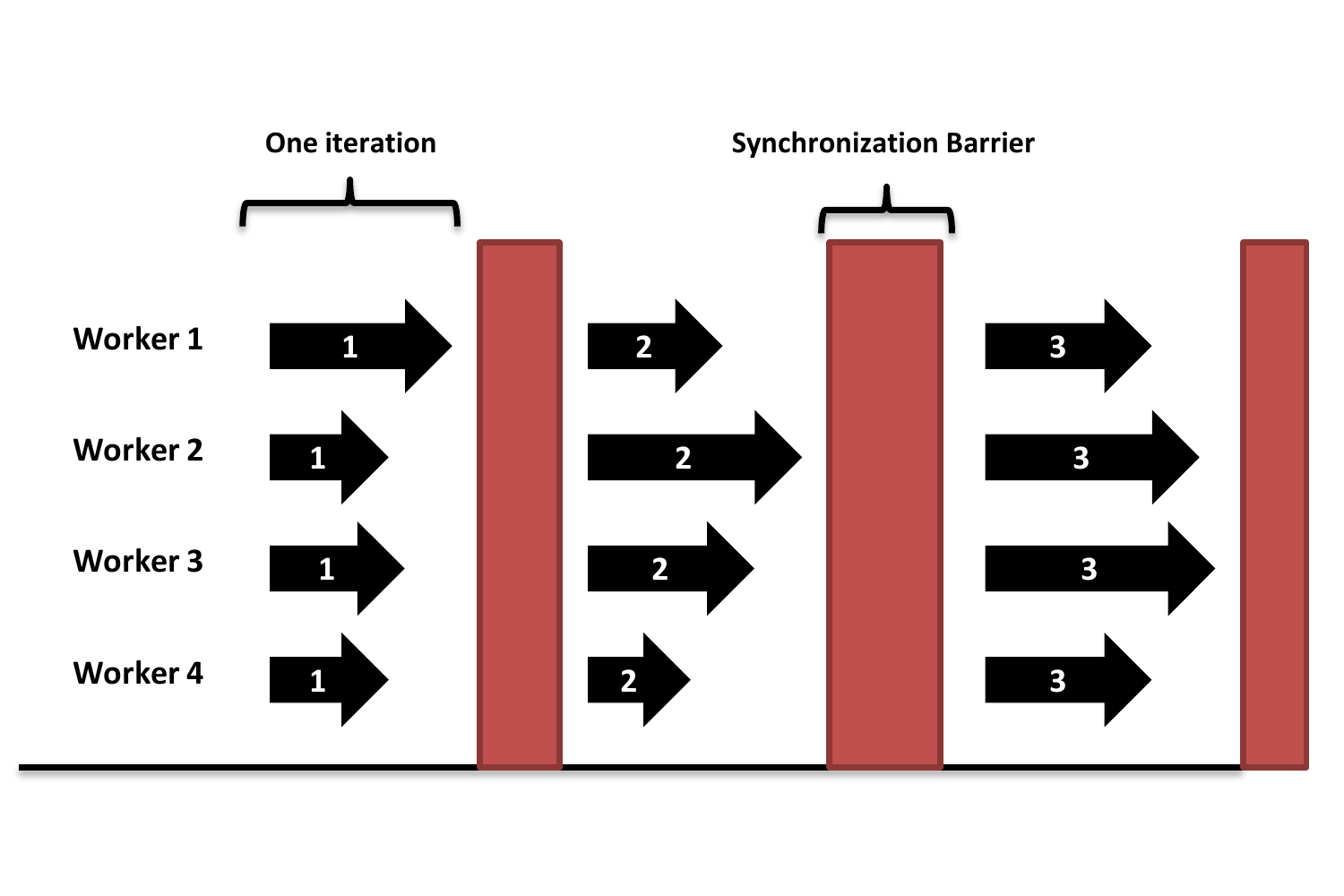}
\caption{Bulk Synchronous Parallel (BSP) Bridging model. For ML programs, the worker machines wait at the end of every iteration for each other, and then exchange information about parameters $\av_j$ during the synchronization barrier.
}
\label{fig:bsp}
\end{center}
\end{figure}

\begin{figure}[t]
\begin{center}
\includegraphics[width=0.6\textwidth]{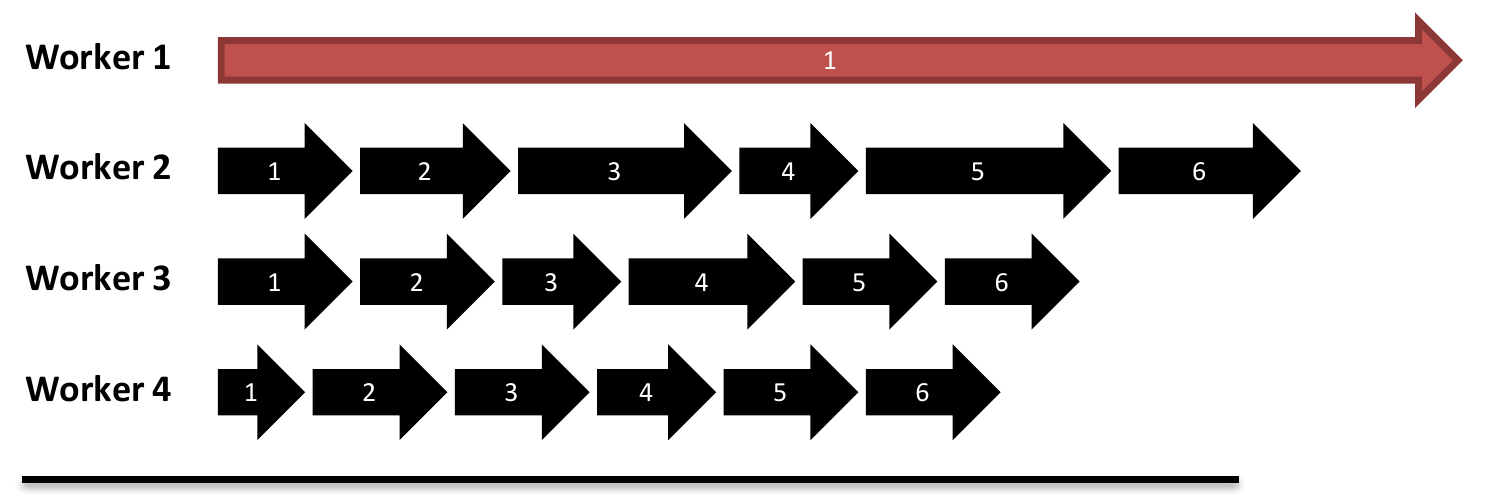}
\caption{Asynchronous Parallel execution. Worker machines running ML programs do not have to wait for each other, and information about model parameters $\av_j$ is exchanged asynchronously and continuously between workers. Because workers do not wait, there is a risk that one machine could end up many iterations slower than the others, which can lead to unrecoverable errors in ML programs. Under a BSP system, this would not happen because of the synchronization barrier.
}
\label{fig:bsp}
\end{center}
\end{figure}

\begin{figure}[t]
\begin{center}
\includegraphics[width=0.6\textwidth]{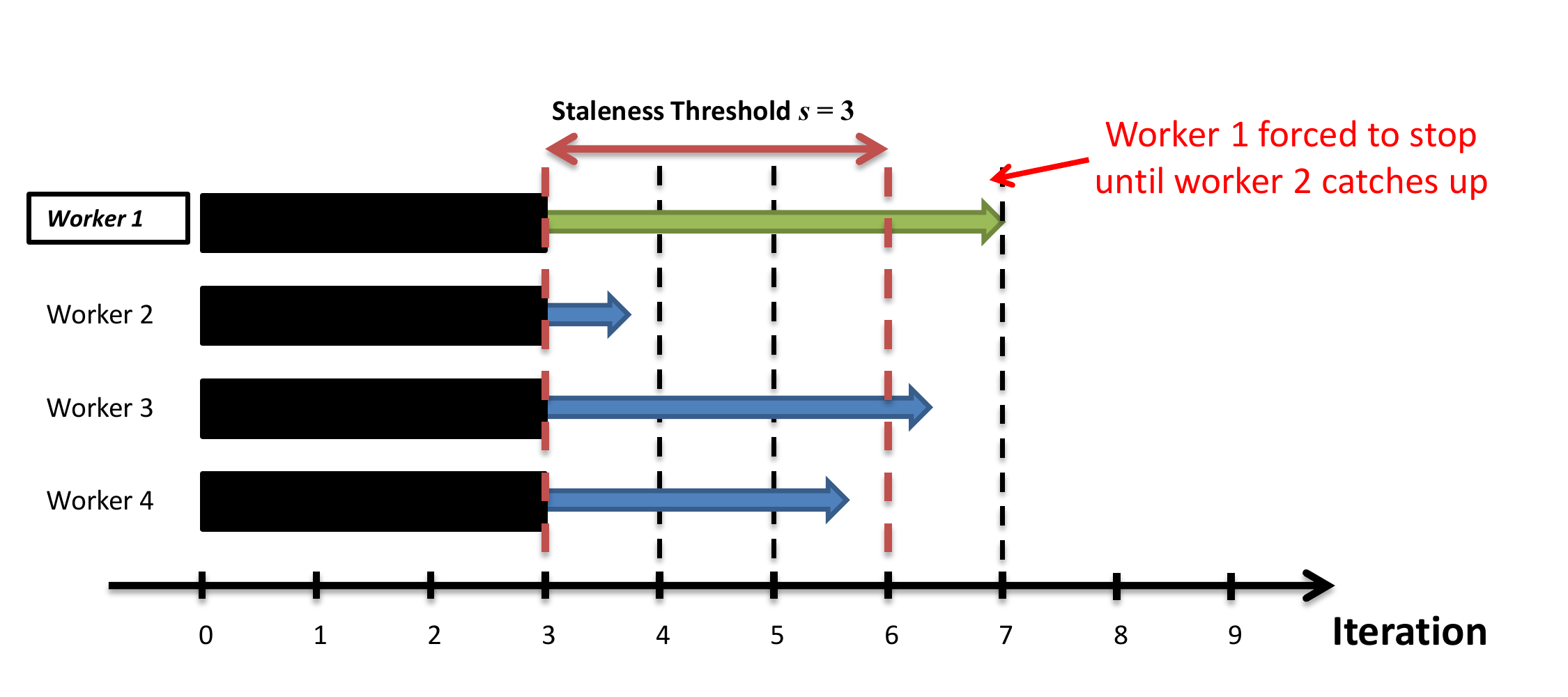}
\caption{Stale Synchronous Parallel (SSP) Bridging model. Compared to BSP, worker machines running ML programs may advance ahead of each other, up to $s$ iterations apart (where $s$ is call the {\it staleness} threshold). Workers that get too far ahead are forced to stop, until slower workers catch up. Like Asynchronous Parallel execution, information about model parameters $\av_j$ is exchanged asynchronously and continuously between workers (with a few additional conditions so as to ensure correct ML convergence), without the need for synchronization barriers. The advantage of SSP is that it behaves like Asynchronous Parallel execution most of the time, yet SSP can also stop workers as needed to ensure correct ML execution.
}
\label{fig:ssp}
\end{center}
\end{figure}

Because BSP creates a clean separation between computation and communication phases, many parallel ML programs running under BSP can be shown to be {\it serializable} --- that is to say, they are equivalent to a sequential ML program. Seralizable BSP ML programs enjoy all the correctness guarantees of their sequential counterparts, and these strong guarantees have made BSP a popular bridging model for both operation-centric programs and ML programs~\cite{dean2008mapreduce,malewicz2010pregel,zaharia2012resilient}. One disadvantage of BSP is that workers must wait for each other to reach the next synchronization barrier, meaning that load-balancing is critical for efficient BSP execution. Yet, even well-balanced workloads can fall prey to {\it stragglers}, machines that become randomly and unpredictably slower than the rest of the cluster~\cite{chilimbi2014project}, due to real-world conditions such as temperature fluctuations in the datacenter, network congestion, and other users' programs or background tasks. When this happens, the program's efficiency drops to match that of the slowest machine (Figure~\ref{fig:bsp}) --- and in a cluster with 1000s of machines, there may even be multiple stragglers. A second disadvantage is that communication between workers is not instantaneous, so the synchronization barrier itself can take a non-trivial amount of time. For example, in LDA topic modeling running on 32 machines under BSP, the synchronization barriers can be up to six times longer than the iterations~\cite{ho2013more}. Due to these two disadvantages, BSP ML programs may suffer from low iteration throughput, i.e. $P$ machines do not produce a $P$-fold increase in throughput.

As an alternative to running ML programs on BSP, asynchronous parallel execution has been explored~\cite{ahmed2012scalable,dean2012large,gonzalez2012powergraph}, in which worker machines never wait for each other, and always communicate model information throughout the course of each iteration. Asynchronous execution usually obtains a near-ideal $P$-fold increase in iteration throughput, but unlike BSP (which ensures serializability and hence ML program correctness), it often suffers from decreased convergence progress per iteration. The reason is that asynchronous communication causes model information to become delayed or {\it stale} (because machines do not wait for each other), and this in turn causes errors in the computation of $\Delta,F$. The magnitude of these errors grows with the delays, and if the delays are not carefully bounded, the result is extremely slow or even incorrect convergence~\cite{ho2013more,dai2015high}. In a sense, there is ``no free lunch" --- model information must be communicated in a timely fashion between workers.

BSP and asynchronous execution face different challenges in achieving ideal $P$-fold ML program speedups --- empirically, BSP ML programs have difficulty reaching the ideal $P$-fold increase in iteration throughput~\cite{ho2013more}, while asynchronous ML programs have difficulty maintaining the ideal progress per iteration observed in sequential ML programs~\cite{ho2013more,dai2015high,zheng2015model}. A promising solution is {\it bounded-asynchronous execution}, in which asychronous execution is permitted up to a limit. To explore this idea further, we present a bridging model called Stale Synchronous Parallel (SSP)~\cite{ho2013more,terry2013replicated}, which generalizes and improves upon BSP.

\noindent
{\bf Stale Synchronous Parallel:}
SSP is a bounded-asynchronous bridging model, which enjoys a similar programming interface to the popular BSP bridging model. An intuitive, high-level explanation goes as follows: we have $P$ parallel workers or machines, that perform ML computations $\Delta,F$ in an iterative fashion. At the end of each iteration $t$, SSP workers signal that they have completed their iterations --- at this point, if the workers were instead running under BSP, a synchronization barrier would be enacted for inter-machine communication. However, SSP does not enact a synchronization barrier. Instead, workers may be stopped or allowed to proceed as SSP sees fit; more specifically, SSP will stop a worker if it is more than $s$ iterations ahead of any other worker, where $s$ is called the {\it staleness threshold} (Figure~\ref{fig:ssp}).

More formally, under SSP, every worker machine keeps an iteration counter $t$, and a {\it local view} of the model parameters $\av$. SSP worker machines ``commit" their updates $\Delta$, and then invoke a {\tt clock()} function that (1) signals that their iteration has ended, (2) increments their iteration counter $t$, (3) informs the SSP system to start communicating $\Delta$ to other machines, so they can update their local views of $\av$. This {\tt clock()} is analogous to BSP's synchronization barrier, but is different in that updates from one worker do not need to be immediately communicated to other workers --- as a consequence, workers may proceeed even if they have only received a partial subset of the updates. This means that the local views of $\av$ can becomes {\it stale}, if some updates have not been received yet.
Given a user-chosen staleness threshold $s \ge 0$, an SSP implementation or system enforces at least the following {\it bounded staleness conditions}:
\begin{itemize}
\item {\bf Bounded clock difference:} The iteration counters on the slowest and fastest workers must be $\leq s$ apart --- otherwise, SSP forces the fastest worker to wait for the slowest worker to catch up.
\item {\bf Timestamped updates:} At the end of each iteration $t$ (right before calling {\tt clock()}), each worker commits an update $\Delta$, which is is timestamped with time $t$.
\item {\bf Model state guarantees:} When a worker with clock $t$ computes $\Delta$, its local view of $\av$ is guaranteed to include all updates $\Delta$ with timestamp $\leq t - s - 1$. The local view may or may not contain updates $\Delta$ from other workers with timestamp $> t - s - 1$.
\item {\bf Read-my-writes:} Each worker will always include its own updates $\Delta$, in its own local view of $\av$.
\end{itemize}
Since the fastest and slowest workers are $ \leq s$ clocks apart, a worker's local view of $\av$ at iteration $t$ will include all updates $\Delta$ from all workers with timestamps in $[0, t - s - 1]$, plus some (or possibly none) of the updates whose timestamps fall in the range $[t - s, t + s - 1]$. Note that SSP is a strict generalization of BSP for ML programs: when $s = 0$, the first range becomes $[0, t - 1]$ while the second range becomes empty, which corresponds exactly to BSP execution of an ML program.

\begin{figure}[t]
\begin{center}
\includegraphics[width=0.8\textwidth]{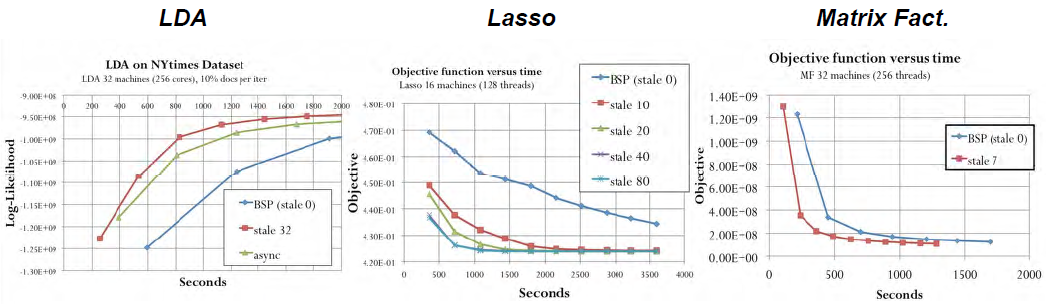}
\caption{(Adapted from~\cite{ho2013more}) Objective function $\obj$ progress versus time plots for three ML programs --- Lasso Regression, Matrix Factorization, Latent Dirichlet Allocation topic modeling --- executed under B\"{o}sen, a system that realizes the Stale Synchronous Parallel (SSP) bridging model. By using SSP (with a range of different staleness values) to improve the iteration throughput of ML algorithms, B\"{o}sen achieves faster time to convergence (steeper curves) than both the BSP bridging model (used in Hadoop and Spark) and fully asynchronous modes of execution. In particular, fully asynchronous execution did not successfully converge for Lasso and Matrix Factorization, and hence the curves are omitted.}
\label{fig:ssp_bosen}
\end{center}
\end{figure}

Because SSP always limits the maximum staleness between any pair of workers to $s$, it enjoys strong theoretical convergence guarantees for both data parallel and model parallel execution. We state two complementary theorems to this effect:
\begin{theorem}[adapted from \cite{dai2015high}]
\label{thm:ssp_data_parallel}
{\bf SSP data parallel Convergence Theorem:}
Consider convex objective functions of the form $\obj = f(\av) = \sum_{t=1}^T f_t(\av)$, where the individual components $f_t$ are also convex. We search for a minimizer $\av^*$ via data parallel stochastic gradient descent on each component $\nabla f_t$ under SSP, with staleness parameter $s$ and $P$ workers. Let the data parallel updates be $\Delta_t:=-\eta_t \nabla_t f_t(\tilde{\av}_t)$ with $\eta_t=\frac{\eta}{\sqrt{t}}$. Under suitable conditions ($f_t$ are $L$-Lipschitz and bounded divergence $D(\av||\av')\le F^2$), we have the following convergence rate guarantee:
\begin{align*}
P\left[\frac{R\left[\av\right]}{T} - \frac{1}{\sqrt{T}} \left(\eta L^{2} + \frac{F^{2}}{\eta} + 2\eta L^2\mu_{\gamma} \right) \ge \tau\right] 
\\
\le \exp\left\{\frac{-T\tau^2}{2\bar{\eta}_T\sigma_{\gamma} + \frac{2}{3}\eta L^2(2s+1)P\tau}\right\}
\end{align*}
where $R[\av] := \sum_{t=1}^T f_t(\tilde{\av}_t) - f(\av^*)$, and $\bar{\eta}_T = \frac{\eta^2 L^4 (\ln T + 1)}{T} = o(1)$ as $T\rightarrow \infty$. In particular, $s$ is the maximum staleness under SSP, $\mu_\gamma$ is the average staleness experienced by the distributed system, and $\sigma_\gamma$ is the variance of the staleness.
\end{theorem}
This data parallel SSP theorem has two implications: first, data parallel execution under SSP is {\it correct} (just like BSP), because $\frac{R[\av]}{T}$ (the difference between the SSP parameter estimate and the true optimum) converges to $\mathcal{O}(T^{-1/2})$ in probability with an exponential tail-bound. Second, it is important to keep the actual staleness and asynchrony as low as possible: the convergence bound becomes tighter with lower maximum staleness $s$, and lower average $\mu_{\gamma}$ and variance $\sigma_{\gamma}$ of the staleness experienced by the workers. For this reason, naive asynchronous systems (e.g. Hogwild!~\cite{hogwild} and YahooLDA~\cite{ahmed2012scalable}) may experience poor convergence in complex production environments, where machines may temporarily slow down due to other tasks or users --- in turn causing the maximum staleness $s$ and staleness variance $\sigma_{\gamma}$ to become arbitrarily large, leading to poor convergence rates.

\begin{theorem}[to appear in 2016]
{\bf SSP model parallel Asymptotic Consistency:}
We consider minimizing objective functions of the form $\obj =  f(\av,D) + g(\av)$ where $\av\in\RR^d$, using a model parallel proximal gradient descent procedure that keeps a centralized ``global view" $\av$ (e.g. on a key-value store) and stale local worker views $\av^p$ on each worker machine.
If the descent step size satisfies $\eta < 1/(L_f+2Ls)$, then the global view $\av$ and local worker views $\av^p$ will satisfy:
\begin{enumerate}
\item\label{sq_summable} $\sum_{t=0}^{\infty} \|\av(t+1) - \av(t)\|^2 < \infty$;
\item\label{sq_vanish} $\lim\limits_{t\to\infty} \|\av(t+1) - \av(t)\| = 0$,  and for all $p$, $\lim\limits_{t\to\infty} \|\av(t) - \av^p(t)\| = 0$;
\item\label{sq_limit} The limit points of $\{\av(t)\}$ coincide with those of $\{\av^p(t)\}$, and both are critical points of $\obj$.
\end{enumerate}
\end{theorem}
Items 1 and 2 imply that the global view $\av$ will eventually stop changing (i.e. converge), and the stale local worker views $\av^p$ will converge to the global view $\av$ --- in other words, SSP model parallel execution will terminate to a stable answer. Item 3 further guarantees that the local and global views $\av^p(t),\av(t)$ will reach an optimal solution to $\obj$ --- in other words, SSP model parallel execution outputs the correct solution. Given additional technical conditions, we can further establish that SSP model parallel execution converges at rate $\mathcal{O}(t^{-1})$.

The above two theorems show that both data parallel and model parallel ML programs running under SSP enjoy near-ideal convergence progress per iteration (that approaches close to BSP and sequential execution).
For example, the B\"{o}sen system~\cite{ho2013more,dai2015high,wei2015managed} uses SSP to achieve up to 10-fold shorter convergence times, compared to the BSP bridging model --- and SSP with properly selected staleness values will not exhibit non-convergence, unlike asynchronous execution (Figure~\ref{fig:ssp_bosen}).
In summary, when SSP is effectively implemented and tuned, it can come close to enjoying the best of both worlds: near-ideal progress per iteration close to BSP, and near-ideal $P$-fold iteration througput similar to asynchronous execution --- and hence, a near-ideal $P$-fold speedup in ML program execution time.

\begin{figure}[t]
\begin{center}
\includegraphics[width=0.6\textwidth]{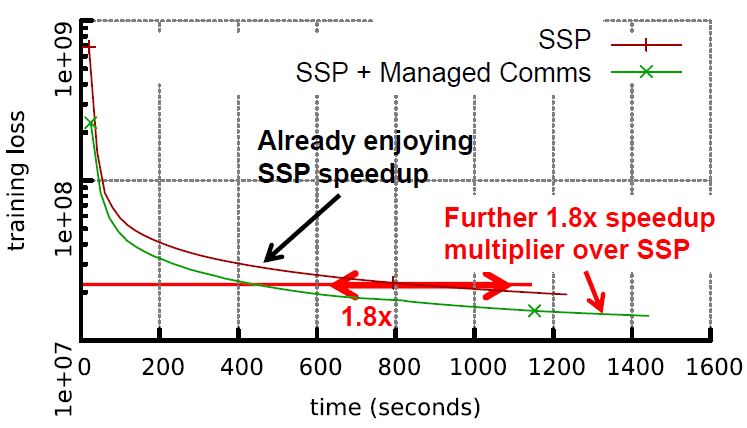}
\caption{(Adapted from~\cite{wei2015managed}) Matrix Factorization: Continuous communication with SSP achieves a further 1.8-times improvement in convergence time over SSP alone. Experiment settings: Netflix dataset with rank 400, on 8 machines (16 cores each) and 1GbE ethernet.
}
\label{fig:mf_bosen}
\end{center}
\end{figure}

\begin{figure}[t]
\begin{center}
\includegraphics[width=0.6\textwidth]{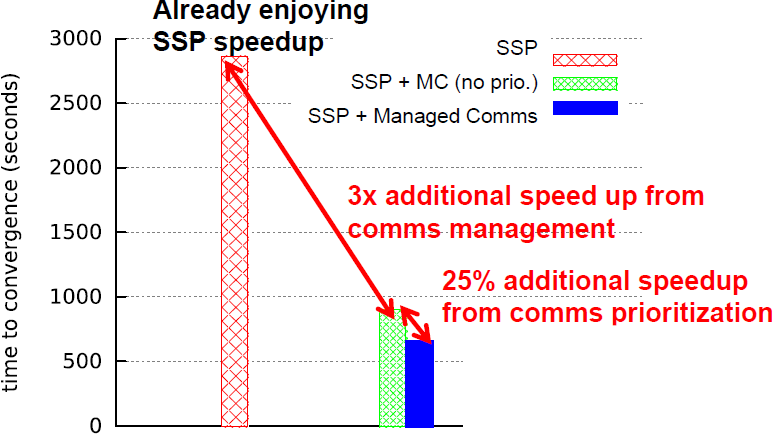}
\caption{(Adapted from~\cite{wei2015managed}) Latent Dirichlet Allocation topic modeling: Continuous communication with SSP achieves a further 3-times improvement in convergence time over SSP alone. Moreover, if update prioritization is also enabled, the convergence time improves by another $25\%$. Experiment settings: NYTimes dataset with 1000 topics, on 16 machines (16 cores each) and 1GbE ethernet.
}
\label{fig:lda_bosen}
\end{center}
\end{figure}

\subsection{
How to Communicate:
Managed Communication and Topologies}
\label{sec:how_to_communicate}

The bridging models (BSP and SSP) just discussed place constraints on when ML computation should occur relative to communication of updates $\Delta$ to model parameters $\av$, in order to guarantee correct ML program execution. However, within the constraints set by a bridging model, there is stil room to prescribe how, or in what order, the updates $\Delta$ should be communicated over the network. Consider the MapReduce sort example, under the BSP bridging model: the Mappers need to send key-value pairs $(a,b)$ with the same key $a$ to the same Reducer. While this can be performed via a {\it bipartite} topology (every Mapper communicates with every Reducer), one could instead use a {\it star} topology, where a third set of machines first aggregates all key-value pairs from the Mappers, and then sends them to the Reducers.

ML algorithms under the SSP bridging model open up an even wider design space --- because SSP only requires updates $\Delta$ to ``arrive no later than $s$ iterations", we could choose to send more important updates first, following the intuition that this should naturally improve algorithm progress per iteration. These considerations are important because every cluster or datacenter's has its own physical switch topology and available bandwidth along each link, and we shall discuss them with the view that choosing the correct {\it communication management} strategy will lead to a noticable improvement in both ML algorithm progress per iteration and iteration throughput. We now discuss several ways in which communication management can be applied to distributed ML systems.

\begin{figure}[t]
\begin{center}
\includegraphics[width=0.6\textwidth]{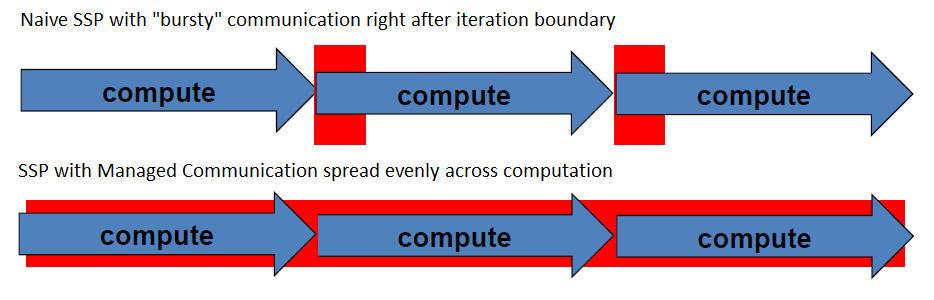}
\caption{Managed Communication in SSP spreads network communication evenly across the duration of computation, instead of sending all updates at once right after the iteration boundary.
}
\label{fig:managed_comm}
\end{center}
\end{figure}

\noindent
{\bf Continuous communication:}
In the first implementations of the SSP bridging model, all inter-machine communication occurred right after the end of each iteration (i.e. right after the SSP \texttt{clock()} command)~\cite{ho2013more}, while leaving the network idle at most other times (Figure~\ref{fig:managed_comm}). The resulting {\it burst} of communication (GBs to TBs) may cause synchronization delays (where updates take longer than expected to reach their destination), and these can be optimized away by adopting a continuous style of communication, where the system waits for existing updates to finish transmission before starting new ones~\cite{wei2015managed}.

Continuous communication can be achieved by a {\it rate limiter} in the SSP implementation, which queues up outgoing communications, and waits for previous communications to finish before sending out the next in line. Importantly, regardless of whether the ML algorithm is data parallel or model parallel, continuous communication still preserves the SSP bounded staleness conditions --- and therefore, it continues to enjoy the same worst-case convergence progress per iteration guarantees as SSP. Furthermore, because managed communication reduces synchronization delays, it also provides a small (2-to-3-fold) speedup to overall convergence time~\cite{wei2015managed}, that is partly due to improved iteration throughput (because of fewer synchronization delays), and partly due to improved progress per iteration (fewer delays also means lower average staleness in local parameter views $\av$, hence SSP's progress per iteration is improved according to Theorem~\ref{thm:ssp_data_parallel}).

\noindent
{\bf Wait-free Back-propagation:}
The deep learning family of ML models~\cite{krizhevsky2012imagenet,dean2012large} presents a special opportunity for continuous communication, due to their highly-layered structure. Two observations stand out in particular: (1) the ``back-propagation" gradient descent algorithm --- used to train deep learning models such as Convolutional Neural Networks (CNNs) --- proceeds in a layer-wise fashion; (2) the layers of a typical CNN (such as ``AlexNet"~\cite{krizhevsky2012imagenet}) are highly asymmetric in terms of model size $|\av|$ and required computation for the back-propagation --- usually, the top fully-connected layers have approximately $90\%$ of the parameters, while the bottom convolutional layers account for $90\%$ of the back-propagation computation~\cite{zhang2015poseidon}. This allows for a specialized type of continuous communication, which we call {\it wait-free back-propagation}: after performing back-propagation on the top layers, the system will communicate their parameters while performing back-propagation on the bottom layers. This spreads the computation and communication out in an optimal fashion, in essence ``overlapping $90\%$ computation with $90\%$ communication".

\noindent
{\bf Update prioritization:}
Another communication management strategy is to {\it prioritize} available bandwidth, by focusing on communicating updates (or parts of) $\Delta$ that contribute most to convergence. This idea has a close relationship with Structure Aware Parallelization discussed in Section~\ref{sec:how_to_distribute} --- while SAP prioritizes computation towards more important parameters, update prioritization ensures that the changes to these important parameters are quickly propagated to other worker machines, so that their effects are immediately felt. As a concrete example, in ML algorithms that use stochastic gradient descent (e.g. Logistic Regression and Lasso Regression), the objective function $\obj$ changes proportionally to the parameters $\av_j$, and hence the fastest-changing parameters $\av_j$ are often the largest contributors to solution quality.

Thus, the SSP implementation can be further augmented by a {\it prioritizer}, which re-arranges the updates in the rate limiter's outgoing queue, so that more important updates will be sent out first.
The prioritizer can support strategies such as the following: (1) Absolute magnitude prioritization: updates to parameters $\av_j$ are re-ordered according to their recent accumulated change $|\delta_j|$, which works well for ML algorithms that use stochastic gradient descent; (2) Relative magnitude prioritization: same as absolute magnitude, but the sorting criteria is $\delta_j/\av_j$, i.e. the accumulated change normalized by the current parameter value $\av_j$.
Empirically, these prioritization strategies already yield another $25\%$ speedup, on top of SSP and continuous communication~\cite{wei2015managed}, and there is potential to explore strategies tailored to a specific ML program (similar to the SAP prioritization criteria for Lasso).

\begin{figure}[t]
\begin{center}
\includegraphics[width=0.6\textwidth]{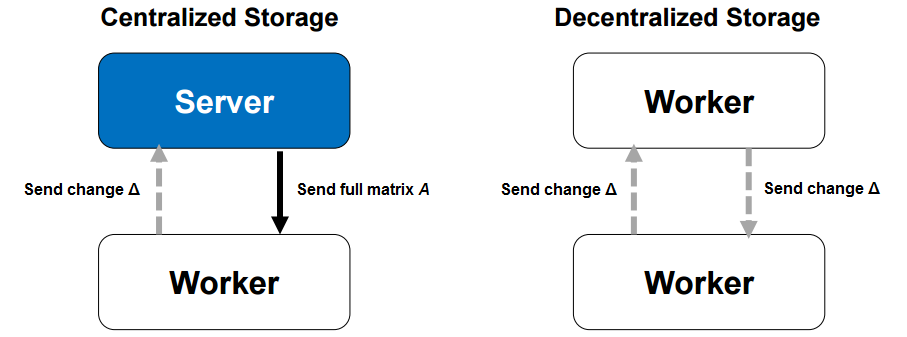}
\caption{Two paradigms for parameter storage: centralized, and decentralized. Note that both paradigms have different communication styles: Centralized storage communicates updates $\Delta$ from workers to the server, and actual parameters $\av$ from servers to workers. Decentralized storage only communicates updates $\Delta$ between workers.
}
\label{fig:model_storage}
\end{center}
\end{figure}

\noindent 
{\bf Parameter Storage and Communication Topologies:}
A third communication management strategy is to consider the placement of model parameters $\av$ across the network ({\it parameter storage}), as well as the network routes along which parameter updates $\Delta$ should be communicated ({\it communication topologies}). The choice of parameter storage strongly influences the communication topologies that can be used, which in turn impacts the speed at which parameter updates $\Delta$ can be delivered over the network (as well as their staleness). Hence, we begin by discussing two commonly-used paradigms for storing model parameters (Fig~\ref{fig:model_storage}): (1) Centralized storage: a ``master view" of the parameters $\av$ is partitioned across a set of server machines, while workers machines maintain local views of the parameters. Communication is asymmetric in the following sense: updates $\Delta$ are sent from the workers to the servers, and workers receive the most up-to-date version of the parameters $\av$ from the server. (2) Decentralized storage: every worker maintains its own local view of the parameters, without a centralized server. Communication is symmetric: workers send updates $\Delta$ to each other, in order to bring their local views of $\av$ up-to-date.

\begin{figure}[t]
\begin{center}
\includegraphics[width=0.6\textwidth]{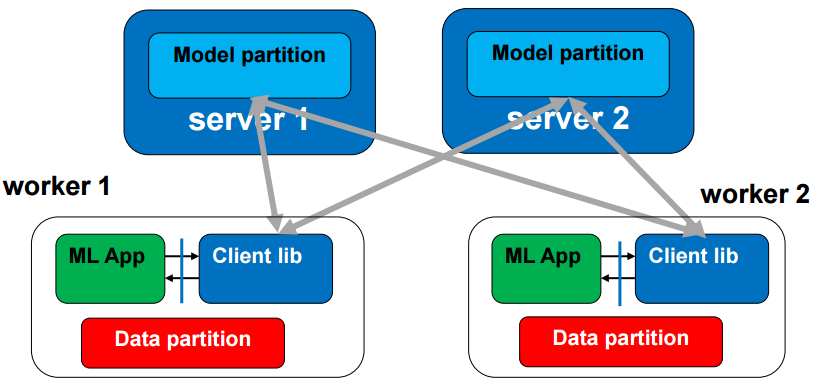}
\caption{Master-slave (bipartite) network topology for centralized parameter storage. Servers only communicate with workers, and vice versa. There is no server-server or worker-worker communciation.
}
\label{fig:ms}
\end{center}
\end{figure}

\begin{figure}[t]
\begin{center}
\includegraphics[width=0.6\textwidth]{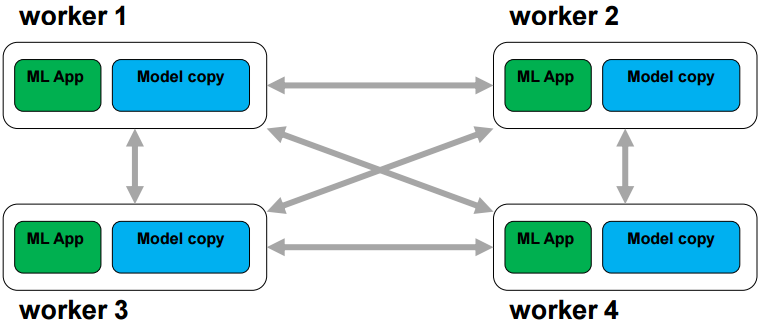}
\caption{Peer-to-peer (P2P) network topology for decentralized parameter storage. All workers may communicate with any other worker.
}
\label{fig:p2p}
\end{center}
\end{figure}

The centralized storage paradigm can be supported by a master-slave network topology (Fig~\ref{fig:ms}), where machines are organized into a bipartite graph with servers on one side, and workers on the other --- whereas
the decentralized storage paradigm can be supported by a peer-to-peer (P2P) topology (Fig~\ref{fig:p2p}), where each worker machine broadcasts to all other workers.
An advantage of the master-slave network topology, is that it reduces the number of messages that need to be sent over the network --- workers only need to send updates $\Delta$ to the servers, which aggregate them using $F$, and update the master view of the parameters $\av$. The updated parameters can then be broadcast to the workers as a single message, rather than a collection of individual updates $\Delta$ --- in total, only $\mathcal{O}(P)$ messages need to be sent. In contrast, P2P topologies must send $\mathcal{O}(P^2)$ messages every iteration, because each worker must broadcast $\Delta$ to every other worker.

\begin{figure}
\begin{center}
\includegraphics[width=0.3\columnwidth]{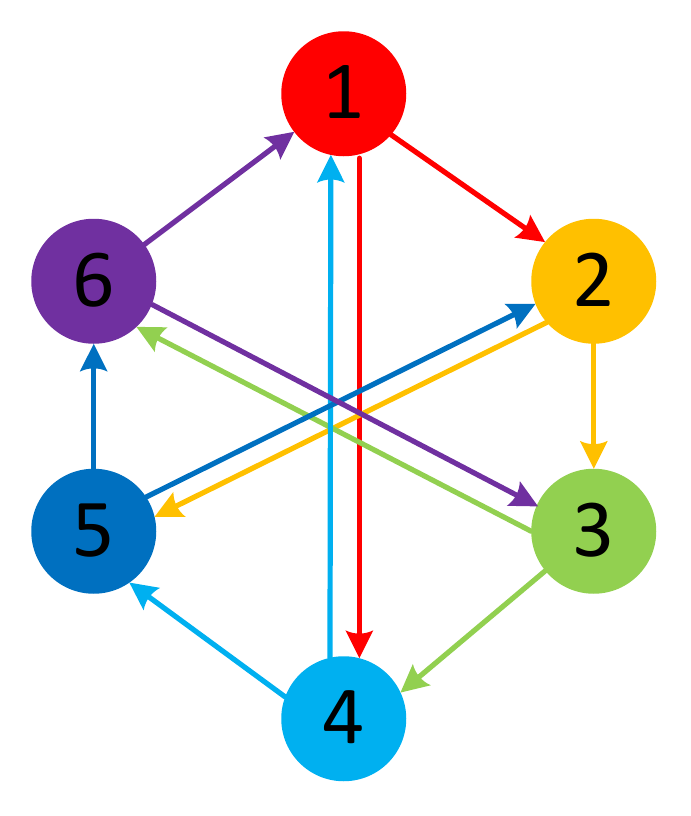}
\caption{Halton Sequence network topology for decentralized parameter storage. Workers may communicate with other workers through an intermediate machine --- for example, worker 1 can reach worker 5 by relaying updates $\Delta$ through worker 2.}
\label{fig:halton}
\end{center}
\vspace{-0.2in}
\end{figure}

However, when $\delta$ has a compact or compressible structure --- such as low-rank-ness in matrix-parameterized ML programs like deep learning, or sparsity in Lasso regression --- the P2P topology can achieve considerable communication savings over the master-slave topology. By compressing or re-representing $\Delta$ in a more compact low-rank or sparse form, each of the $\mathcal{O}(P^2)$ P2P messages can be made much smaller than the $\mathcal{O}(P)$ master-to-slave messages, which may not admit compression (because the messages consist of the actual parameters $\av$, not the compressible updates $\Delta$).
Furthermore, even the $\mathcal{O}(P^2)$ P2P messages can be reduced, by switching from a full P2P to a partially-connected Halton Sequence topology (Fig~\ref{fig:halton})~\cite{li2015malt}, where each worker only communicates with a subset of workers. Workers can reach any other worker by routing messages through intermediate nodes: for example, the routing path $1\to2\to5\to6$ is one way to send a message from worker 1 to 6. The intermediate nodes can combine messages meant for the same destination, thus reducing the number of messages per iteration (and further reducing network load). However, one drawback to the Halton Sequence topology is that routing increases the time taken for messages to reach their destination, which raises the average staleness of parameters under the SSP bridging model --- e.g. the message from worker 1 to 6 would be three iterations stale. The Halton Sequence topology is nevertheless a good option for very large cluster networks, which have limited peer-to-peer bandwidth.

By combining the various aspects of ``how to communicate" --- continuous communication, update prioritization, and a suitable combination of parameter storage and communication topology ---  we can design a distributed ML system that enjoys multiplicative speed benefits from each aspect, resulting in an almost-order-of-magnitude speed improvement on top of what SAP (how to distribute) and SSP (bridging models) can offer. For example, the B\"{o}sen SSP system enjoys up to an additional 4-fold speedup from continuous communication and update prioritization, as shown in Figure~\ref{fig:mf_bosen} and~\ref{fig:lda_bosen}~\cite{wei2015managed}.

\subsection{What to Communicate}

Going beyond how to store and communicate updates $\Delta$ between worker machines, we may also ask ``what" needs to be communicated in each update $\Delta$ --- in particular, is there any way to reduce the number of bytes required to transmit $\Delta$, and thus further alleviate the comunication bottleneck in distribute ML programs~\cite{xie2015distributed}? This question is related to the idea of lossless compression in operation-centric programs; for example, Hadoop Mapreduce is able to compresses key-value pairs $(a,b)$ to reduce their transmission cost from Mappers to Reducers. For data parallel ML programs, a commonly-used strategy for reducing the size of $\Delta$ messages is to aggregate (i.e. sum) them before transmission over the network, taking advantage of the additive structure within $F$ (such as in the Lasso data parallel example, Eq~\ref{eq:data_parallel_lasso}). Such early aggregation is preferred for centralized parameter storage paradigms that communicate full parameters $\av$ from servers to workers~\cite{ho2013more,dai2015high}, and it is natural to ask if there are other strategies, that may perhaps be better-suited to different storage paradigms.

To answer this question, we may inspect the mathematical structure of ML parameters $\av$, and the nature of their updates $\Delta$. A number of popular ML programs have matrix-structured parameters $\mb{\av}$ (we use boldface to distinguish from the generic $\av$) --- examples include multiclass logistic regression (MLR), neural networks (NN) \cite{chilimbi2014project}, distance metric learning (DML)~\cite{xing2002distance} and sparse coding~\cite{olshausen1997sparse}.
We refer to these as {\it matrix-parameterized models} (MPMs), and note that $\mb{\av}$ can be very large in modern applications: in one application of MLR to Wikipedia~\cite{partalas2015lshtc}, $\mb{\av}$ is a $325$k-by-$10$k matrix containing several billion entries (10s of GBs). It is also worth pointing out that typical computer cluster networks can at most transmit a few GBs per second between two machines, hence naive synchronization of such matrices $\mb{\av}$ and their updates $\Delta$ is not instantaneous. Because parameter synchronization occurs many times across the lifetime of an iterative-convergent ML program, the time required for synchronization can become a substantial bottleneck. 

More formally, an MPM is an ML objective functions with the following specialized form:
\begin{align}
\label{eq:mpm}
\obj(\mb{x},\mb{\av}) &= \min_{\mb{\av}} \left[ \frac{1}{N}\sum_{i=1}^{N}f_i(\mb{\av}\mb{u}_i, \mb{v}_i) \right] + r(\mb{\av})
\end{align} 
where the model parameters are a $K$-by-$D$ matrix $\mb{\av}\in \RR^{K\times D}$, and each loss function $f_i$ is defined over $\mb{\av}$ and the data samples $\mb{x} = \{(\mb{u}_i,\mb{v}_i)\}_{i=1}^{N}$ --- specifically, $f_i$ must depend on the product $\mb{\av}\mb{u}_i$ (and not $\mb{\av}$ or $\mb{u}_i$ individually). $r(\mb{\av})$ is a structure-inducing function such as a regularizer.
A well-known example of Eq.~\ref{eq:mpm} is Multiclass logistic regression (MLR), which is used in classification problems involving tens of thousands of classes $K$ (e.g. web data collections like Wikipedia). In MLR, $\mb{\av}$ is the weight coefficient matrix, $\mb{u}_i$ is the $D$-dimensional feature vector of data sample $i$, $\mb{v}_i$ is a $K$-dimensional indicator vector representing the class label of data sample $i$, and the loss function $f_i$ is composed of a cross-entropy error function and a softmax mapping of $\mb{\av}\mb{u}_i$. A key property of MPMs is that each update $\Delta$ is a {\it low-rank} matrix, and can be factored into small vectors, called {\it sufficient factors}, that are cheap to transmit over the network.

\noindent
{\bf Sufficient Factor Broadcasting:}
In order to exploit the sufficient factor property in MPMs, let us look closely at the updates $\Delta$. The ML objective function Eq.~\ref{eq:mpm} can be solved by either the stochastic proximal gradient descent (SPGD)~\cite{dean2012large,ho2013more,chilimbi2014project,li2015malt} or stochastic dual coordinate ascent (SDCA)~\cite{hsieh2008dual,shalev2013stochastic,yang2013trading,jaggi2014communication,hsieh2015comm} algorithmic techniques, amongst others.
For example, in SPGD, the update function $\Delta$ can be decomposed into a sum over vectors $\mb{b}_i\mb{c}_i^\top$, where $\mb{b}_i=\frac{\partial f(\mb{\av}\mb{u}_i,\mb{v}_i)}{\partial (\mb{\av}\mb{u}_i)}$ and $\mb{c}_i=\mb{u}_{i}$; SDCA updates $\Delta$ also admit a similar decomposition\footnote{More generally, $\mb{b}_i,\mb{c}_i$ may be ``thin matrices" instead of vectors. Sufficient Factor Broadcasting works as long as $\mb{b}_i,\mb{c}_i$ are much smaller than $\mb{\av}$.}~\cite{xie2015distributed}. Instead of transmitting $\Delta = \sum_i \mb{b}_i\mb{c}_i^\top$ (total size $KD$) between workers, we can instead transmit the individual vectors $\mb{b}_i,\mb{c}_i$ (total size $S(K+D)$, where $S$ is the number of data samples processed in the current iteration), and reconstruct $\Delta$ at the destination machine.

This {\it sufficient factor broadcasting} strategy is well-suited to decentralized storage paradigms, where only updates $\Delta$ are transmitted between workers. It may also be applied to centralized storage paradigms, though only for transmissions from workers to servers; the server-to-worker direction sends full matrices $\mb{\av}$ that no longer have the sufficient factor property~\cite{chilimbi2014project}. At this point, it is natural to ask how the combination of decentralized storage and sufficient factor broadcasting interacts with the SSP bridging model --- will the ML algorithm still output the correct answer under such a P2P setting? The following theorem provides an affirmative answer:
\begin{theorem}[adapted from~\cite{xie2015distributed}]
{\bf Sufficient Factor Broadcasting under SSP, Convergence Theorem:}
\label{thm:model_2}
Let $\mb{\av}_p(t)$, $p=1, \ldots, P$, and $\mb{\av}(t)$ be the local worker views and a ``reference" view respectively, for the ML objective function $\obj$ in Eq.~\ref{eq:mpm} (assuming $r \equiv 0$) being solved by sufficient factor broadcasting under the SSP bridging model with staleness $s$. Under mild assumptions,
we have
\begin{enumerate}
\item $\lim\limits_{t\to\infty} \max_p \|\mb{\av}(t) - \mb{\av}_p(t)\| = 0$, i.e. the local worker views converge to the reference view, implying that all worker views will be the same after sufficient iterations $t$.
\item There exists a common subsequence of $\mb{\av}_p(t)$ and $\mb{\av}(t)$ that converges almost surely to a stationary point of $\obj$, with rate
$\mathcal{O} \left(\frac{P s\log(t)}{\sqrt{t}}\right)$
\end{enumerate}
\end{theorem}
Intuitively, Theorem~\ref{thm:model_2} says that
all local worker views $\mb{\av}_p(t)$ eventually converge to stationary points (local minima) of the objective function $\obj$, even though updates $\Delta$ can be stale by up to $s$ iterations. Thus, sufficient factor broadcasting under decentralized storage is robust under the SSP bridging model --- which is especially useful for topologies like Halton Sequence that increase the staleness of updates, in exchange for lower bandwidth usage.

\begin{figure}[t]
\begin{center}
\includegraphics[width=0.24\columnwidth]{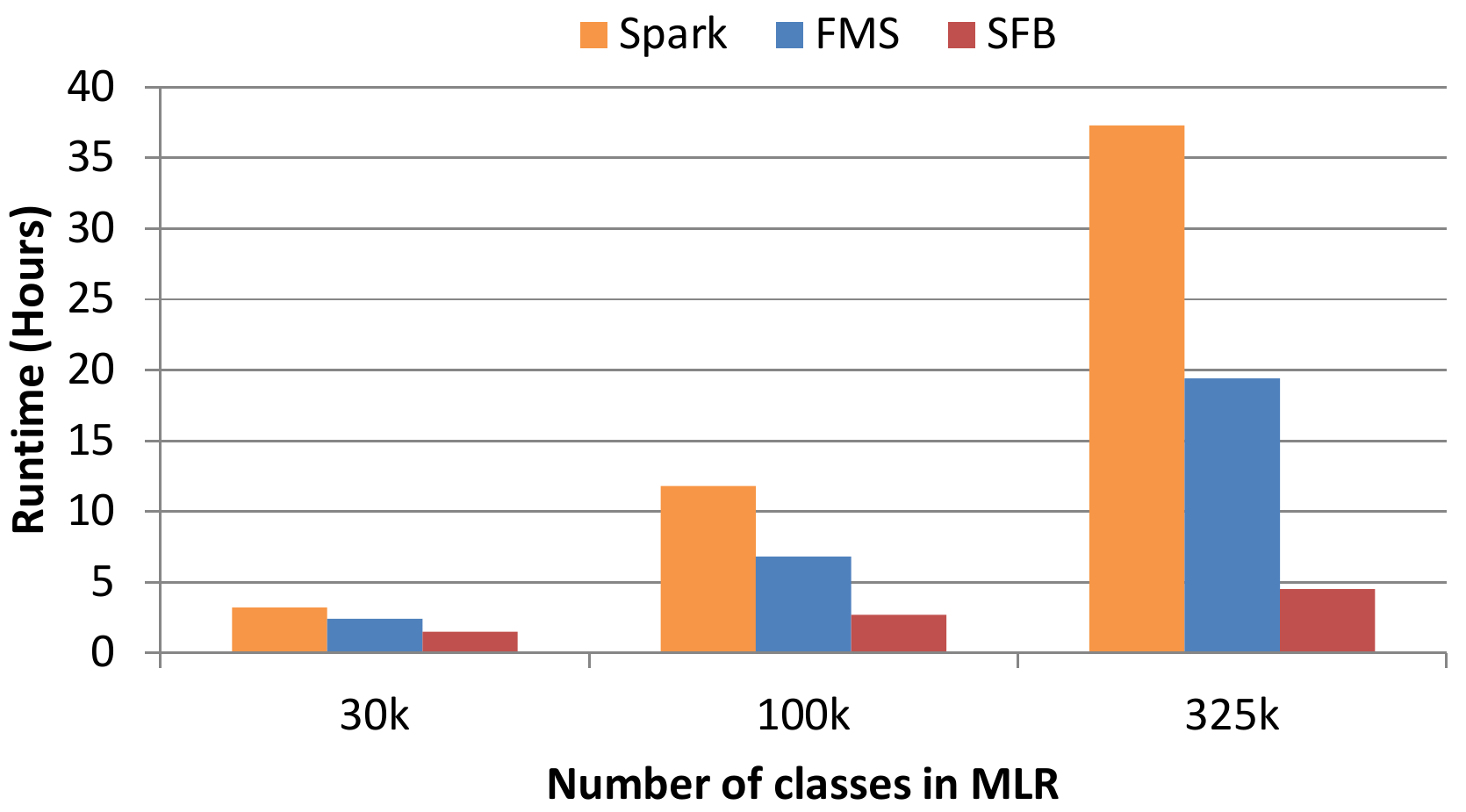}
\includegraphics[width=0.24\columnwidth]{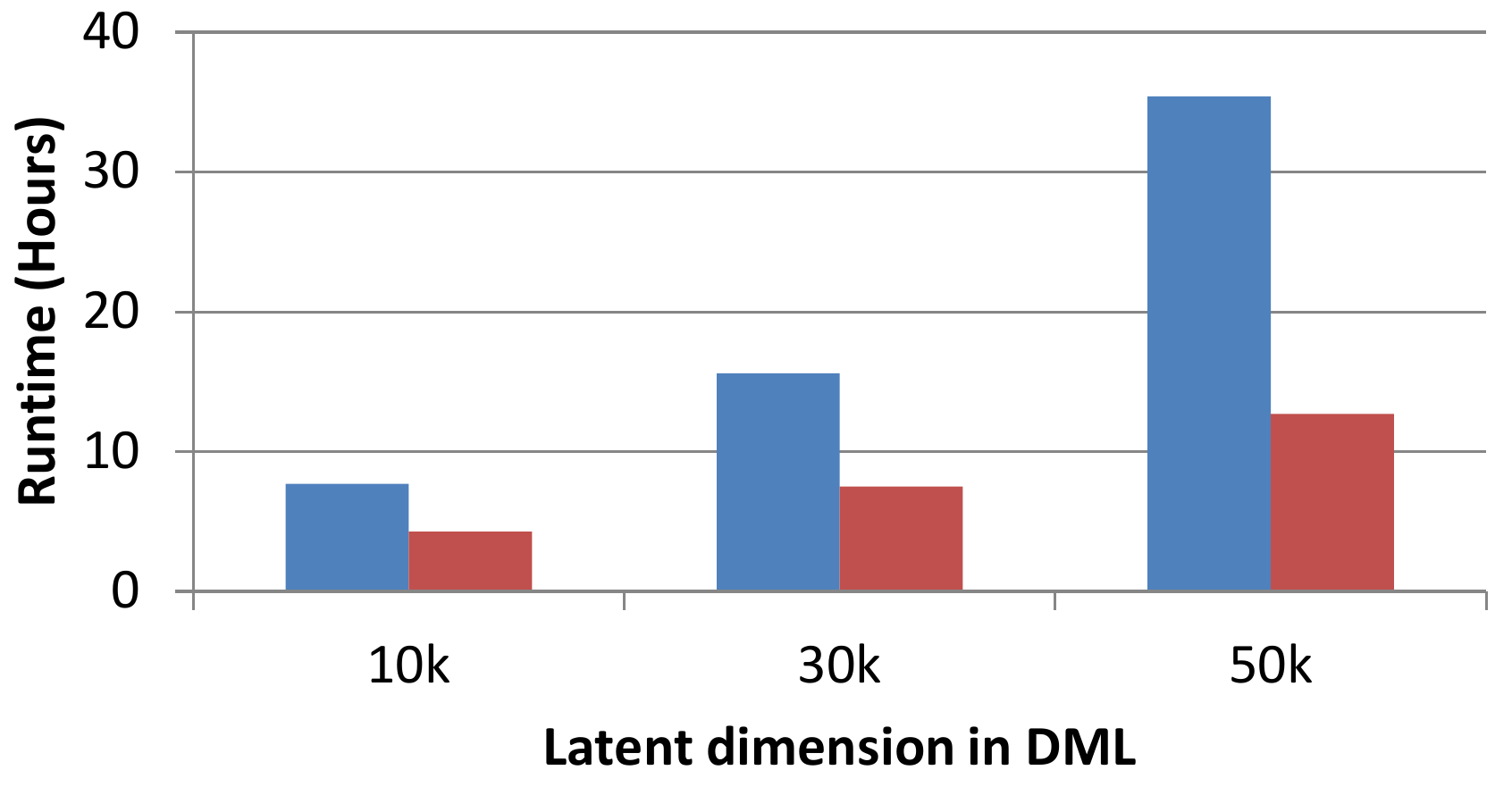}
\includegraphics[width=0.24\columnwidth]{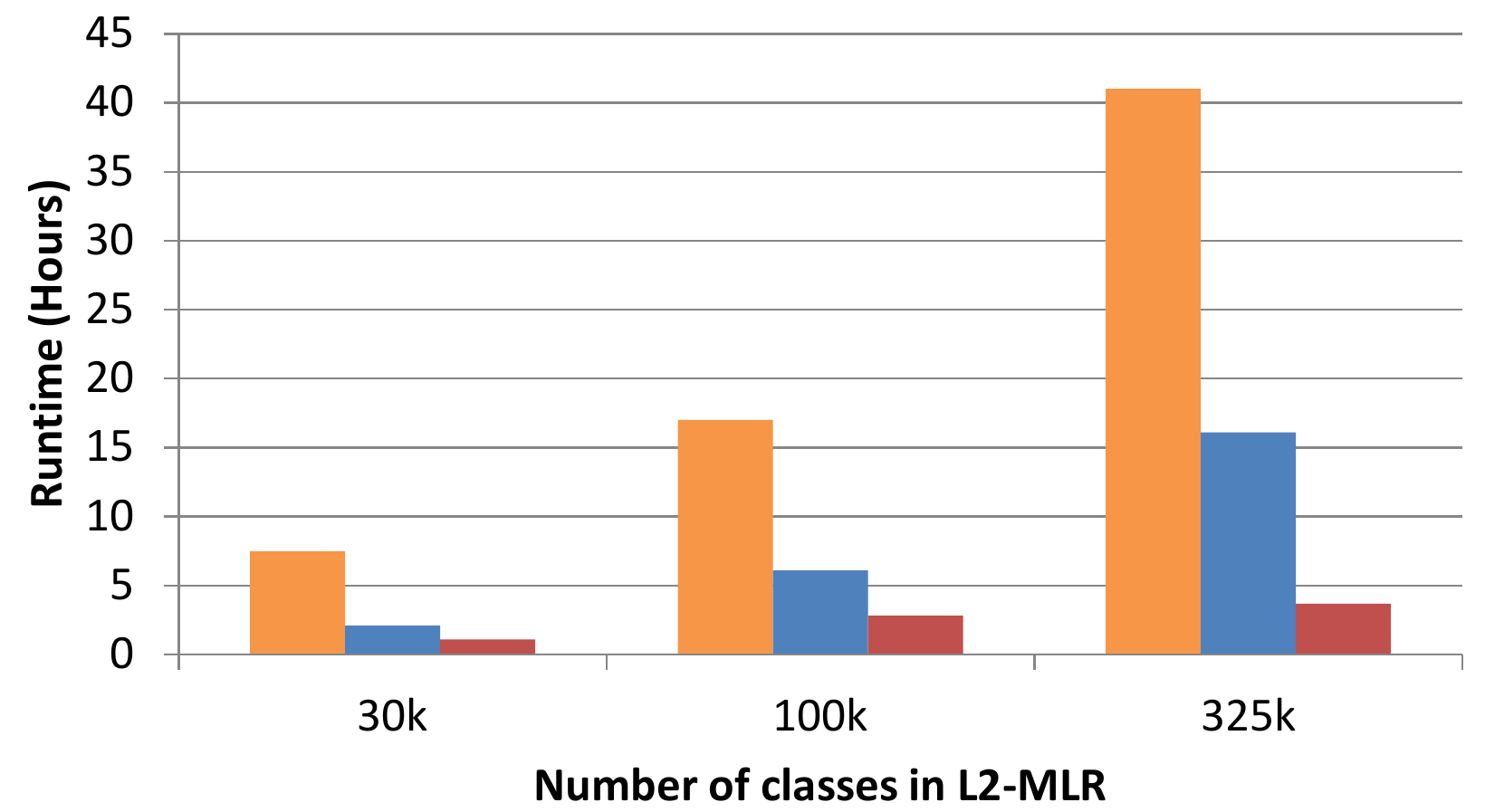}
\caption{Convergence time versus model size for MLR, DML, L2-MLR (left to right).}
\label{fig:exp_runtime}
\end{center}
\end{figure}

\begin{figure}[t]
\begin{center}
\includegraphics[width=0.3\columnwidth]{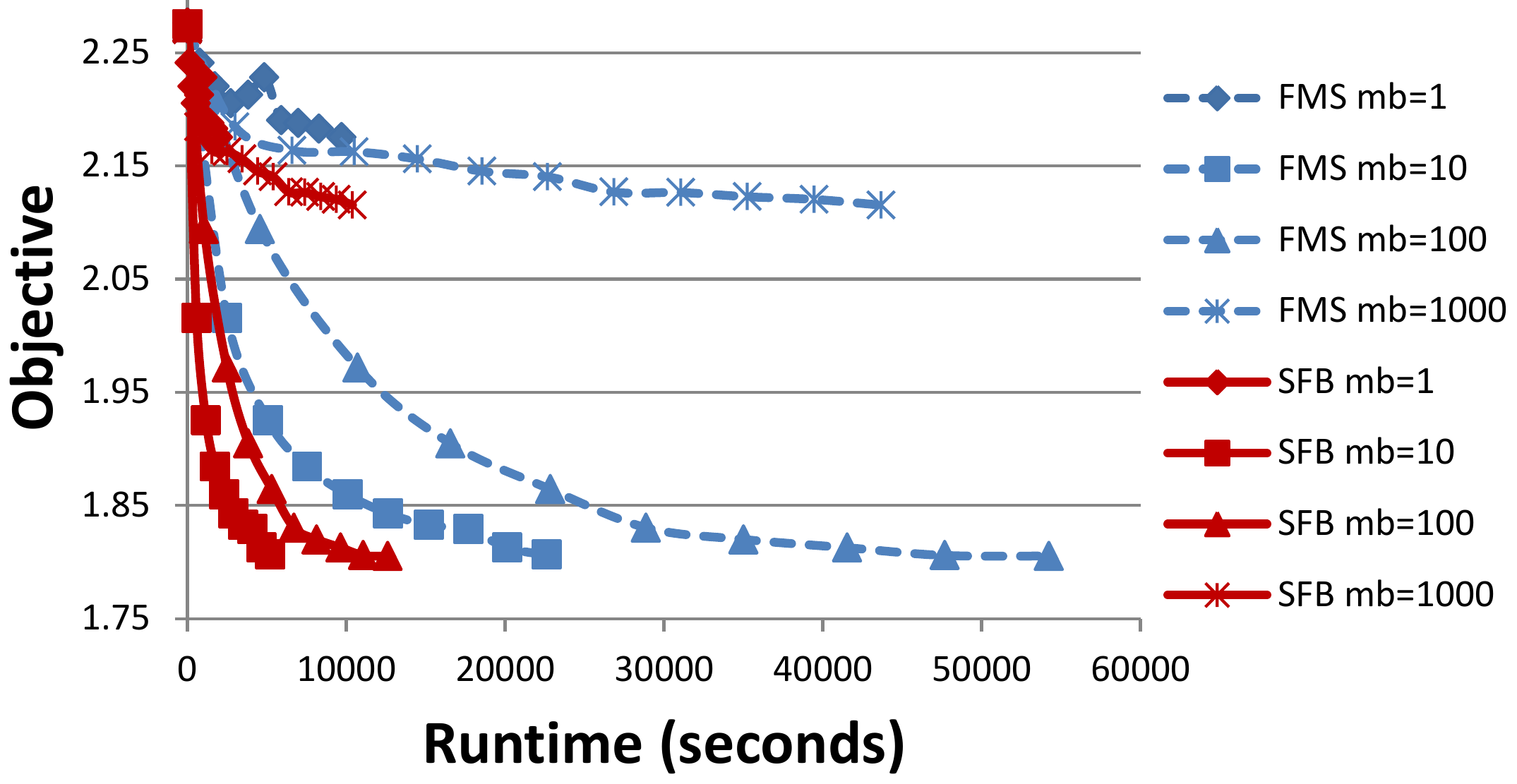}
\includegraphics[width=0.3\columnwidth]{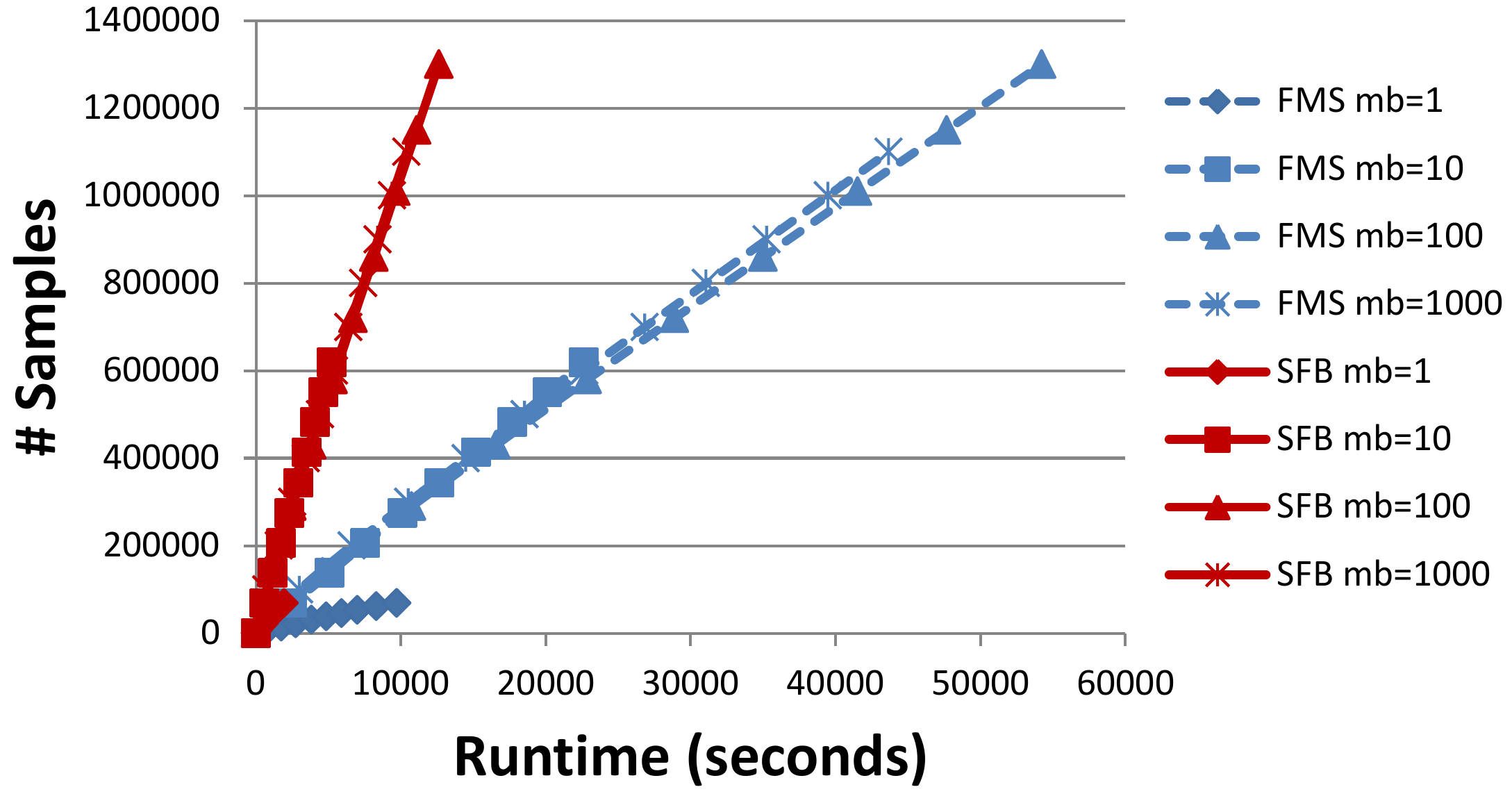}
\includegraphics[width=0.3\columnwidth]{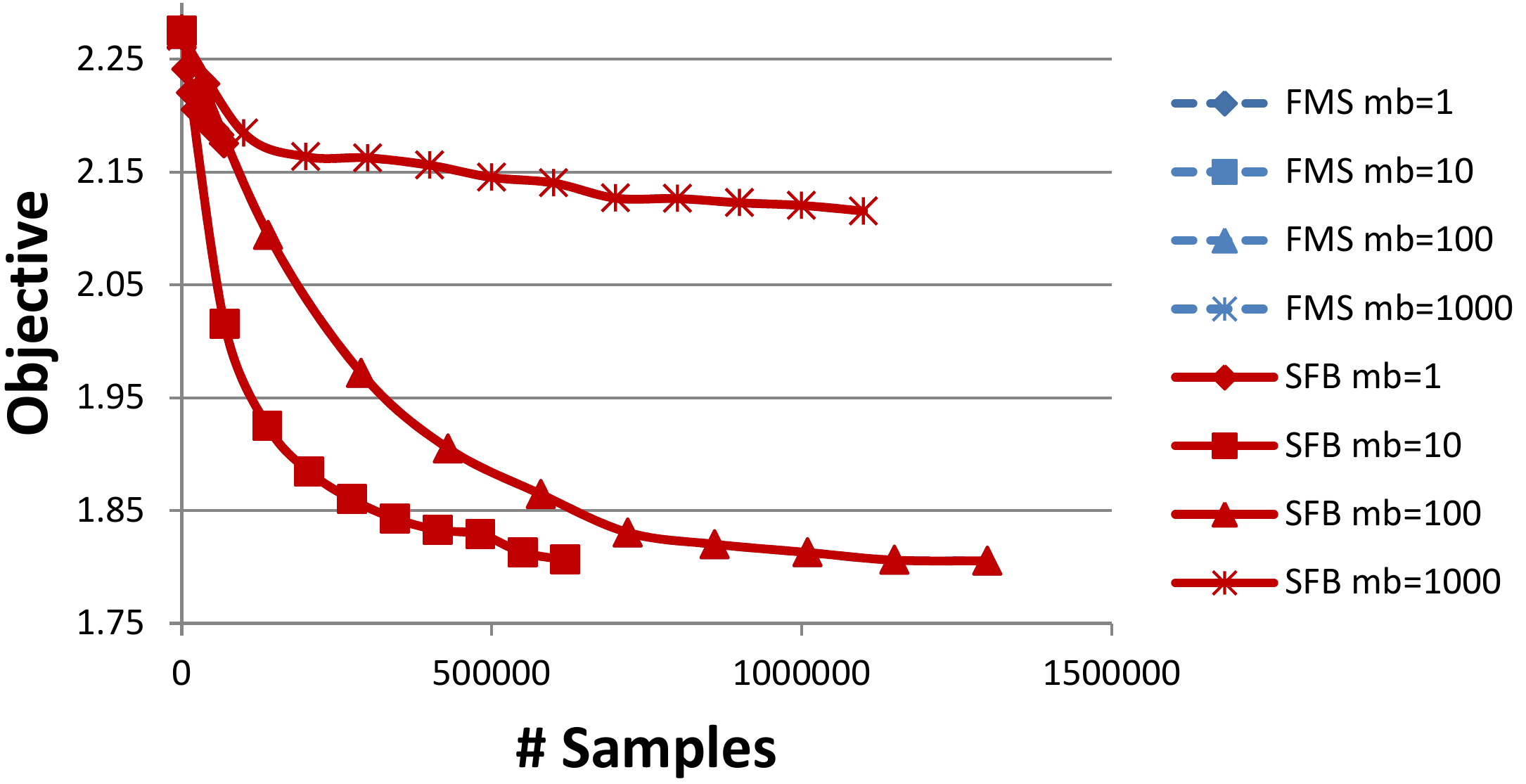}
\caption{MLR objective vs runtime (left), samples vs runtime (middle),  objective vs samples (right).}
\label{fig:iter_qtt_qlt}
\end{center}
\end{figure}

\begin{figure}[t]
\begin{center}
\includegraphics[width=0.24\columnwidth]{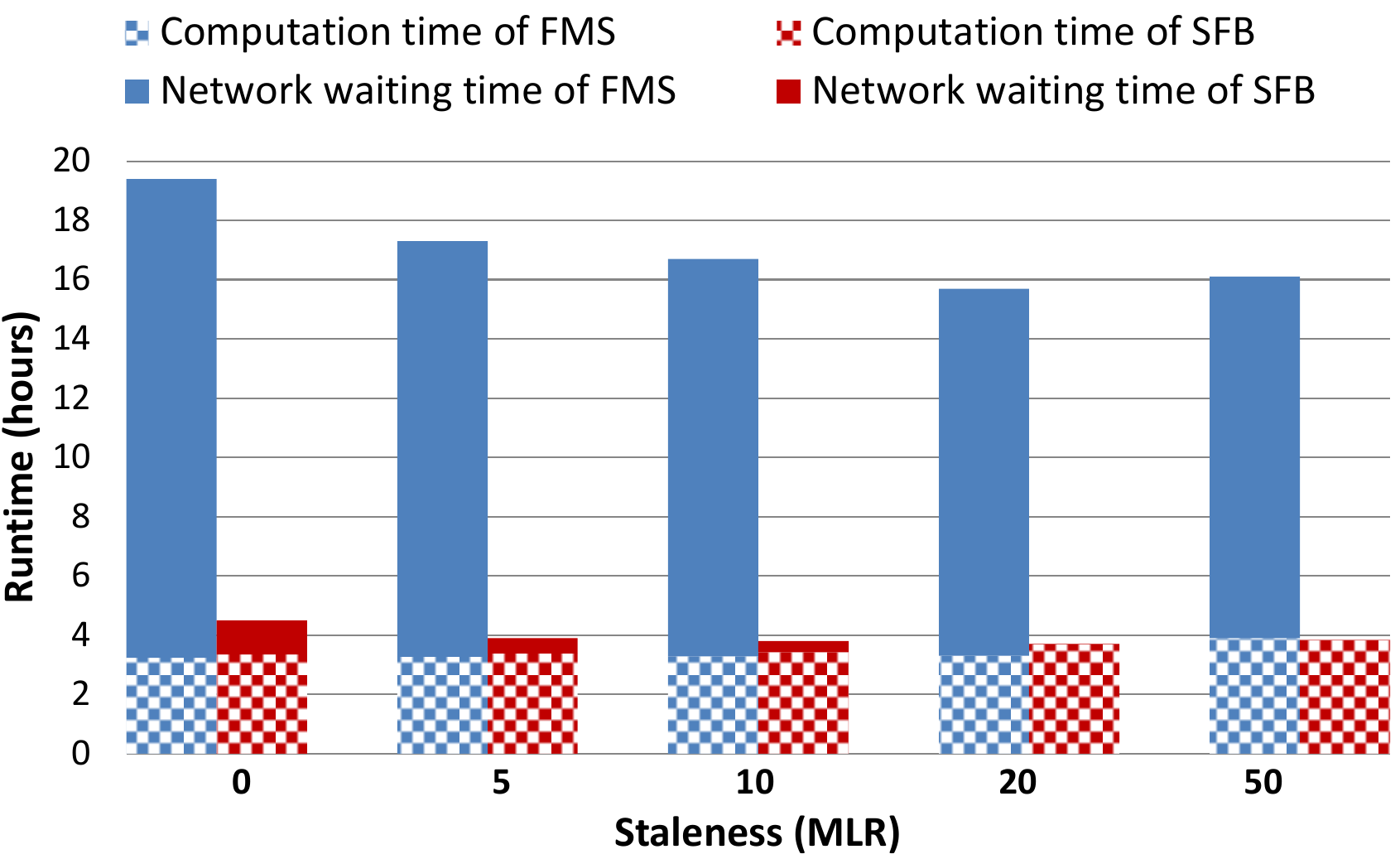}
\includegraphics[width=0.24\columnwidth]{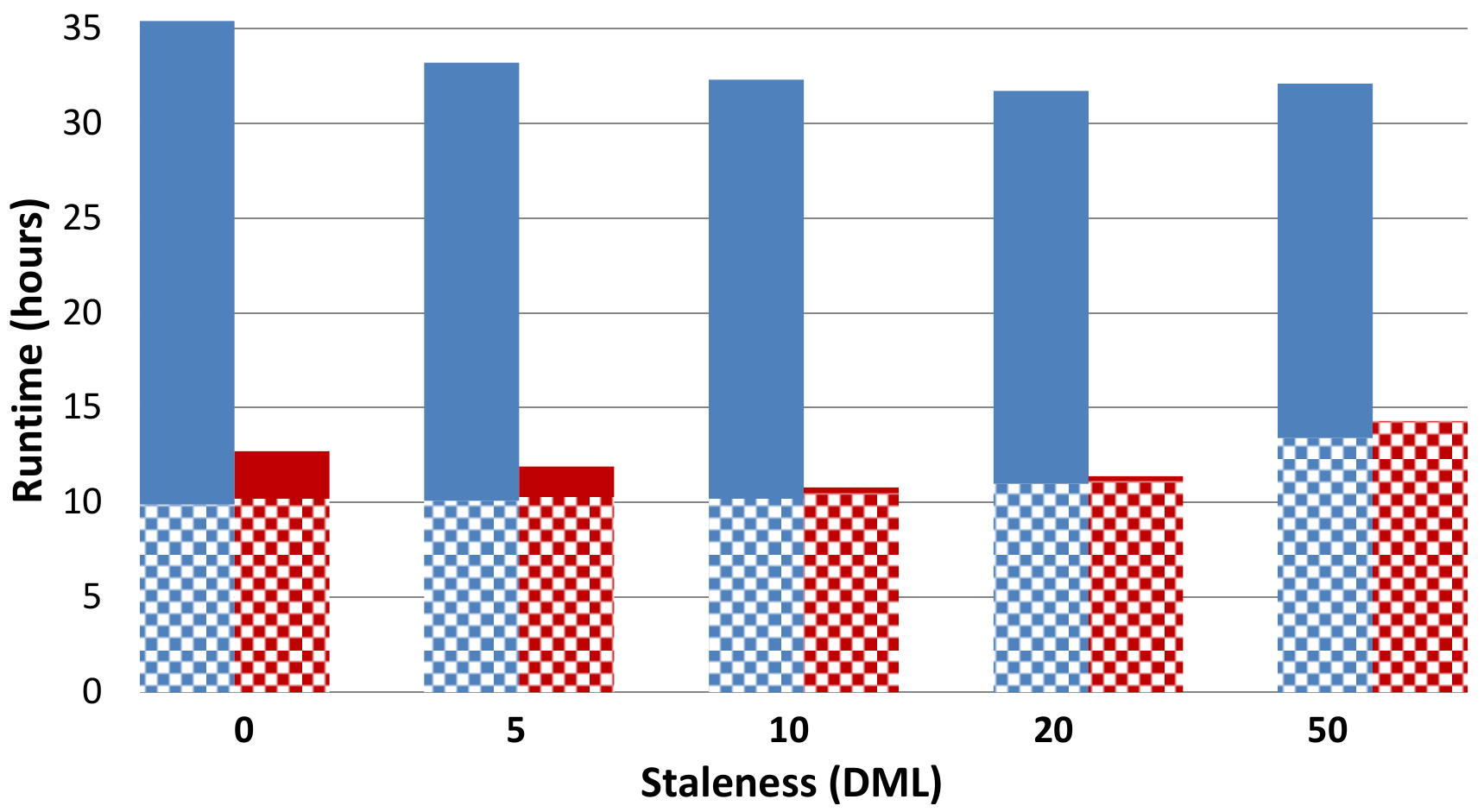}
\includegraphics[width=0.24\columnwidth]{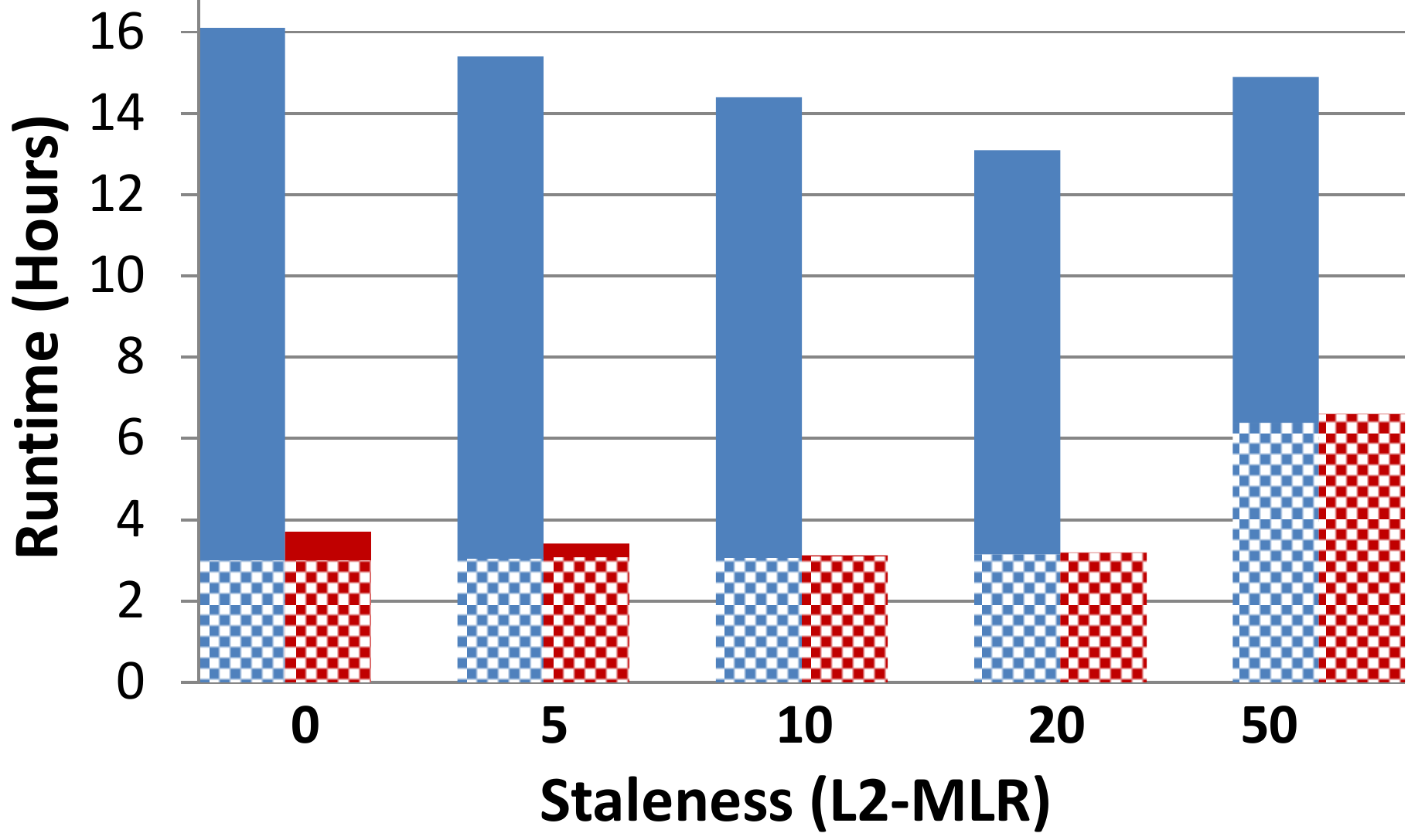}
\caption{Computation vs network waiting time for MLR, DML, L2-MLR (left to right).}
\label{fig:exp_timebreak}
\end{center}
\end{figure}

Empirically, sufficient factor broadcasting (SFB) can greatly reduce the communication costs for matrix parametrized models (MPMs): for a variety of MPMs, Figure \ref{fig:exp_runtime} shows the time taken to reach a fixed objective value using the BSP bridging model.
MPMs running under SFB converge faster than when running under a centralized storage paradigm that transmits full updates $\Delta$ (referred to as ``full matrix synchronization" or FMS); we also compare to baselines implementations included with Spark v1.3.1 (not all MPM being evaluated are available on Spark).
This is because SFB has lower communication costs, so a greater proportion of algorithm running time is spent on computation instead of network waiting; we show this in Figure \ref{fig:iter_qtt_qlt}, which plots data samples processed per second
(i.e. iteration throughput) and algorithm progress per sample (i.e. progress per iteration) for multiclass logistic regression (MLR), under BSP consistency and varying minibatch sizes. The middle graph shows that SFB processes far more samples per second than FMS, while the rightmost graph shows that SFB and FMS yield exactly the same algorithm progress per sample under BSP.

To understand the impact of SFB on $\Delta$ communication costs, let us examine Figure \ref{fig:exp_timebreak}, which shows the {\it total} computation time as well as network communication time required by SFB and FMS to converge, across a range of SSP staleness values
--- in general, higher $\Delta$ communication costs and lower staleness will increase the time the ML program spends waiting for network communication.
For all staleness values, SFB requires far less network waiting (because SFs are much smaller than full matrices in FMS).
Computation time for SFB is slightly longer than FMS because (1) update matrices $\Delta$ must be reconstructed on each SFB worker, and (2) SFB requires a few more iterations for convergence than FMS, due to slightly higher average parameter staleness compared to FMS.
Overall, SFB's reduction in network waiting time far surpasses the added computation time, and hence SFB outperforms FMS.

As a final note, there are situations that naturally call for a hybrid of sufficient factor broadcasting and full $\Delta$ transmission. A good example is deep learning using Convolutional Neural Networks (previously discussed under the topic of wait-free back-propagation in Section~\ref{sec:how_to_communicate}): the top layers of a typical CNN are fully-connected, and use matrix parameters containing millions of elements, whereas the bottom layers are convolutional and involve tiny matrices with at most a few hundred elements. It follows that it is more efficient to (1) apply sufficient factor broadcasting to the top layers' updates (transmission cost is $S(K+D) \ll KD$ because $K,D$ are large relative to $S$); (2) aggregate (sum) the bottom layers' updates before transmission (cost is $KD \ll S(K+D)$ because $S$ is large relative to $K,D$)~\cite{zhang2015poseidon}.

%% file: conclusion.tex
\section{Petuum: a Realization of the ML System Design Principles}
\begin{figure}
\centering
\includegraphics[width=0.6\textwidth]{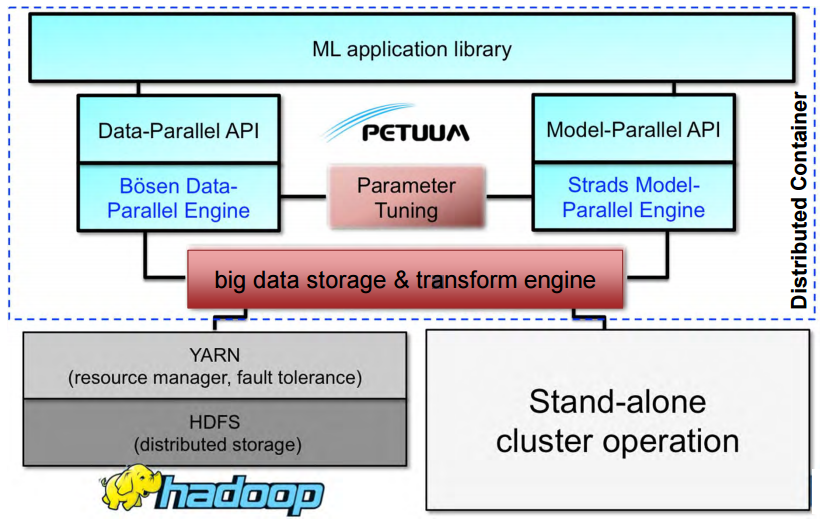}
\caption{Architecture of Petuum, a distributed ML system for Big Data and Big Models.}
\label{fig:petuum}
\end{figure}

We conclude this paper by noting that the four principles of ML system design have been partially realized by systems that are highly-specialized for one or a few ML programs~\cite{li2014scaling,ahmed2012scalable,chilimbi2014project,hogwild,feature_cluster}. This presents ML practitioners with a choice between the aforementioned monolithic yet high-performance ``towers" (specialized systems that require substantial engineering to maintain and upgrade), or the more general-purpose yet slower ``platforms" such as Hadoop and Spark (which are relatively easy to deploy and maintain). In order to address this dichotomy, we have realized the principles of ML system design in the Petuum distributed machine learning framework~\cite{xing2015petuum}, whose architecture is outlined in Figure~\ref{fig:petuum}. The intent behind Petuum is to provide a generic distributed systems for ML algorithms running on Big Data, by abstracting system implementation details and the four design principles away from the ML programmer --- who is then freed to focus on programming the key ML functions $\obj,\Delta,F$.

Compared to general-purpose distributed programming platforms for operation-centric programs (such as Hadoop and Spark), Petuum takes advantage of the unique properties of iterative-convergence ML programs --- error tolerance, dependency structures, non-uniform convergence and compact updates --- in order to improve both the convergence rate and per-iteration time for ML algorithms, and thus achieve close to ideal $P$-fold speedup with $P$ machines. Petuum runs on compute clusters and cloud compute, supporting from 10s to 1000s of machines, and provides programming interfaces for C++ and Java, while also supporting YARN and HDFS to allow execution on existing Hadoop clusters. Two major systems underlie Petuum (Figure~\ref{fig:petuum}): (1) B\"{o}sen, a bounded-asynchronous distributed key-value store for data-parallel ML programming. B\"{o}sen uses the Stale Synchronous Parallel consistency model, which allows asynchronous-like performance that outperforms MapReduce and bulk synchronous execution, yet does not sacrifice ML algorithm correctness; (2) Strads, a dynamic scheduler for model-parallel ML programming. Strads performs fine-grained scheduling of ML update operations, prioritizing computation on the parts of the ML program that need it most, while avoiding unsafe parallel operations that could lead to non-convergence in ML programs.

Currently, Petuum features an ML library with over 10 ready-to-run algorithms (implemented on top of Bösen and Strads), including classic algorithms such as logistic regression, k-means, and random forest and newer algorithms such as supervised topic models (MedLDA), deep learning, distance metric learning and sparse coding. In particular, the Petuum deep learning system, Poseidon~\cite{zhang2015poseidon}, fully exemplifies the ``platform" nature of Petuum: Poseidon takes the well-established but single-machine Caffe project\footnote{\url{http://caffe.berkeleyvision.org/}}, and turns it into a distributed GPU system by replacing the memory access routines within Caffe with the B\"{o}sen distributed key-value store's Distributed Shared Programming programming interfaces. The biggest advantage of this platform approach is familiarity and usability --- existing Caffe users do not have to learn a new tool in order to take advantage of GPUs distributed across a cluster.

Looking towards the future, we envision that Petuum might become the foundation of a {\it ML Distributed Cluster Operating System} that provides a single-machine or laptop-like experience for ML application users and programmers, while making full use of the computational capacity provided by datacenter-scale clusters with 1000s of machines. Achieving this vision will certainly require new systems such as containerization, cluster resource management and scheduling, and user interfaces to be developed, which are necessary steps to reduce the substantial human or {\it operational cost} of deploying massive-scale ML applications in a datacenter environment. By building such systems into the ML-centric Petuum platform --- which reduces the {\it capital cost} of ML applications by enabling them to run faster on fewer machines --- we can thus prepare for the eventual Big Data computational shift from database-style operations to ML-style operations.

%% file: main.bbl
\begin{thebibliography}{10}

\bibitem{agarwal2011distributed}
A.~Agarwal and J.~C. Duchi.
\newblock Distributed delayed stochastic optimization.
\newblock In {\em NIPS}, 2011.

\bibitem{ahmed2012scalable}
A.~Ahmed, M.~Aly, J.~Gonzalez, S.~Narayanamurthy, and A.~J. Smola.
\newblock Scalable inference in latent variable models.
\newblock In {\em WSDM}, 2012.

\bibitem{ahmed2011unified}
A.~Ahmed, Q.~Ho, J.~Eisenstein, E.~Xing, A.~J. Smola, and C.~H. Teo.
\newblock Unified analysis of streaming news.
\newblock In {\em Proceedings of the 20th international conference on World
  wide web}, pages 267--276. ACM, 2011.

\bibitem{airoldi2008mixed}
E.~M. Airoldi, D.~M. Blei, S.~E. Fienberg, and E.~P. Xing.
\newblock Mixed membership stochastic blockmodels.
\newblock {\em Journal of Machine Learning Research}, 9:1981--2014, 2008.

\bibitem{blei2003latent}
D.~M. Blei, A.~Ng, and M.~Jordan.
\newblock Latent dirichlet allocation.
\newblock {\em JMLR}, 3:993--1022, 2003.

\bibitem{bottou2010large}
L.~Bottou.
\newblock Large-scale machine learning with stochastic gradient descent.
\newblock In {\em Proceedings of COMPSTAT'2010}, pages 177--186. Springer,
  2010.

\bibitem{shotgun}
J.~K. Bradley, A.~Kyrola, D.~Bickson, and C.~Guestrin.
\newblock Parallel coordinate descent for l1-regularized loss minimization.
\newblock In {\em ICML}, 2011.

\bibitem{burges1998tutorial}
C.~J. Burges.
\newblock A tutorial on support vector machines for pattern recognition.
\newblock {\em Data mining and knowledge discovery}, 2(2):121--167, 1998.

\bibitem{chandola2009anomaly}
V.~Chandola, A.~Banerjee, and V.~Kumar.
\newblock Anomaly detection: A survey.
\newblock {\em ACM computing surveys (CSUR)}, 41(3):15, 2009.

\bibitem{chilimbi2014project}
T.~Chilimbi, Y.~Suzue, J.~Apacible, and K.~Kalyanaraman.
\newblock Project adam: building an efficient and scalable deep learning
  training system.
\newblock In {\em OSDI}, 2014.

\bibitem{coates2013deep}
A.~Coates, B.~Huval, T.~Wang, D.~Wu, B.~Catanzaro, and N.~Andrew.
\newblock Deep learning with cots hpc systems.
\newblock In {\em Proceedings of the 30th international conference on machine
  learning}, pages 1337--1345, 2013.

\bibitem{dai2015high}
W.~Dai, A.~Kumar, J.~Wei, Q.~Ho, G.~Gibson, and E.~P. Xing.
\newblock High-performance distributed ml at scale through parameter server
  consistency models.
\newblock In {\em AAAI}. 2015.

\bibitem{dean2012large}
J.~Dean, G.~Corrado, R.~Monga, K.~Chen, M.~Devin, M.~Mao, A.~Senior, P.~Tucker,
  K.~Yang, Q.~V. Le, et~al.
\newblock Large scale distributed deep networks.
\newblock In {\em NIPS}, 2012.

\bibitem{mapreduce03}
J.~Dean and S.~Ghemawat.
\newblock Mapreduce: Simplified data processing on large clusters.
\newblock In {\em OSDI 2004}, pages 137--150, 2004.

\bibitem{dean2008mapreduce}
J.~Dean and S.~Ghemawat.
\newblock Mapreduce: simplified data processing on large clusters.
\newblock {\em Communications of the ACM}, 51(1):107--113, 2008.

\bibitem{fercoq2013accelerated}
O.~Fercoq and P.~Richt{\'a}rik.
\newblock Accelerated, parallel and proximal coordinate descent.
\newblock {\em arXiv preprint arXiv:1312.5799}, 2013.

\bibitem{gemulla2011large}
R.~Gemulla, E.~Nijkamp, P.~J. Haas, and Y.~Sismanis.
\newblock Large-scale matrix factorization with distributed stochastic gradient
  descent.
\newblock In {\em Proceedings of the 17th ACM SIGKDD international conference
  on Knowledge discovery and data mining}, pages 69--77. ACM, 2011.

\bibitem{ghahramani2005infinite}
Z.~Ghahramani and T.~L. Griffiths.
\newblock Infinite latent feature models and the indian buffet process.
\newblock In {\em Advances in neural information processing systems}, pages
  475--482, 2005.

\bibitem{gilks2005markov}
W.~R. Gilks.
\newblock {\em Markov chain monte carlo}.
\newblock Wiley Online Library, 2005.

\bibitem{gonzalez2012powergraph}
J.~E. Gonzalez, Y.~Low, H.~Gu, D.~Bickson, and C.~Guestrin.
\newblock Powergraph: distributed graph-parallel computation on natural graphs.
\newblock In {\em OSDI}, 2012.

\bibitem{hinton2012deep}
G.~Hinton, L.~Deng, D.~Yu, G.~E. Dahl, A.-r. Mohamed, N.~Jaitly, A.~Senior,
  V.~Vanhoucke, P.~Nguyen, T.~N. Sainath, et~al.
\newblock Deep neural networks for acoustic modeling in speech recognition: The
  shared views of four research groups.
\newblock {\em Signal Processing Magazine, IEEE}, 29(6):82--97, 2012.

\bibitem{hinton2006reducing}
G.~E. Hinton and R.~R. Salakhutdinov.
\newblock Reducing the dimensionality of data with neural networks.
\newblock {\em Science}, 313(5786):504--507, 2006.

\bibitem{ho2013more}
Q.~Ho, J.~Cipar, H.~Cui, S.~Lee, J.~K. Kim, P.~B. Gibbons, G.~A. Gibson,
  G.~Ganger, and E.~Xing.
\newblock More effective distributed ml via a stale synchronous parallel
  parameter server.
\newblock In {\em NIPS}, 2013.

\bibitem{hsieh2008dual}
C.-J. Hsieh, K.-W. Chang, C.-J. Lin, S.~S. Keerthi, and S.~Sundararajan.
\newblock A dual coordinate descent method for large-scale linear svm.
\newblock In {\em ICML}, 2008.

\bibitem{hsieh2015comm}
C.-J. Hsieh, H.-F. Yu, and I.~S. Dhillon.
\newblock Passcode: Parallel asynchronous stochastic dual co-ordinate descent.
\newblock In {\em ICML}, 2015.

\bibitem{jaggi2014communication}
M.~Jaggi, V.~Smith, M.~Tak{\'a}c, J.~Terhorst, S.~Krishnan, T.~Hofmann, and
  M.~I. Jordan.
\newblock Communication-efficient distributed dual coordinate ascent.
\newblock In {\em NIPS}, 2014.

\bibitem{kim2012tree}
S.~Kim, E.~P. Xing, et~al.
\newblock Tree-guided group lasso for multi-response regression with structured
  sparsity, with an application to eqtl mapping.
\newblock {\em The Annals of Applied Statistics}, 6(3):1095--1117, 2012.

\bibitem{koller2009probabilistic}
D.~Koller and N.~Friedman.
\newblock {\em Probabilistic graphical models: principles and techniques}.
\newblock MIT press, 2009.

\bibitem{krizhevsky2012imagenet}
A.~Krizhevsky, I.~Sutskever, and G.~E. Hinton.
\newblock Imagenet classification with deep convolutional neural networks.
\newblock In {\em Advances in neural information processing systems}, pages
  1097--1105, 2012.

\bibitem{kumar2014fugue}
A.~Kumar, A.~Beutel, Q.~Ho, and E.~P. Xing.
\newblock Fugue: Slow-worker-agnostic distributed learning for big models on
  big data.
\newblock In {\em AISTATS}, 2014.

\bibitem{lee1999learning}
D.~D. Lee and H.~S. Seung.
\newblock Learning the parts of objects by non-negative matrix factorization.
\newblock {\em Nature}, 1999.

\bibitem{lee2006efficient}
H.~Lee, A.~Battle, R.~Raina, and A.~Y. Ng.
\newblock Efficient sparse coding algorithms.
\newblock In {\em Advances in neural information processing systems}, pages
  801--808, 2006.

\bibitem{lee2014model}
S.~Lee, J.~K. Kim, X.~Zheng, Q.~Ho, G.~Gibson, and E.~P. Xing.
\newblock On model parallelism and scheduling strategies for distributed
  machine learning.
\newblock In {\em NIPS}. 2014.

\bibitem{lee2012leveraging}
S.~Lee and E.~P. Xing.
\newblock Leveraging input and output structures for joint mapping of epistatic
  and marginal eqtls.
\newblock {\em Bioinformatics}, 28(12):i137--i146, 2012.

\bibitem{li2015malt}
H.~Li, A.~Kadav, E.~Kruus, and C.~Ungureanu.
\newblock Malt: distributed data-parallelism for existing ml applications.
\newblock In {\em Proceedings of the Tenth European Conference on Computer
  Systems}, 2015.

\bibitem{li2014scaling}
M.~Li, D.~G. Andersen, J.~W. Park, A.~J. Smola, A.~Ahmed, V.~Josifovski,
  J.~Long, E.~J. Shekita, and B.-Y. Su.
\newblock Scaling distributed machine learning with the parameter server.
\newblock In {\em OSDI}, 2014.

\bibitem{graphlab10}
Y.~Low, J.~Gonzalez, A.~Kyrola, D.~Bickson, C.~Guestrin, and J.~M. Hellerstein.
\newblock Graphlab: A new parallel framework for machine learning.
\newblock In {\em UAI}, Catalina Island, California, 2010.

\bibitem{graphlab12}
Y.~Low, J.~Gonzalez, A.~Kyrola, D.~Bickson, C.~Guestrin, and J.~M. Hellerstein.
\newblock {Distributed GraphLab: A Framework for Machine Learning and Data
  Mining in the Cloud}.
\newblock {\em PVLDB}, 2012.

\bibitem{malewicz2010pregel}
G.~Malewicz, M.~H. Austern, A.~J. Bik, J.~C. Dehnert, I.~Horn, N.~Leiser, and
  G.~Czajkowski.
\newblock Pregel: a system for large-scale graph processing.
\newblock In {\em SIGMOD}, 2010.

\bibitem{mccoll1995bulk}
W.~F. McColl.
\newblock Bulk synchronous parallel computing.
\newblock In {\em Abstract machine models for highly parallel computers}, pages
  41--63. Oxford University Press, 1995.

\bibitem{mnih2007probabilistic}
A.~Mnih and R.~Salakhutdinov.
\newblock Probabilistic matrix factorization.
\newblock In {\em Advances in neural information processing systems}, pages
  1257--1264, 2007.

\bibitem{moritz2015sparknet}
P.~Moritz, R.~Nishihara, I.~Stoica, and M.~I. Jordan.
\newblock Sparknet: Training deep networks in spark.
\newblock {\em arXiv preprint arXiv:1511.06051}, 2015.

\bibitem{hogwild}
F.~Niu, B.~Recht, C.~R{\'e}, and S.~J. Wright.
\newblock Hogwild!: A lock-free approach to parallelizing stochastic gradient
  descent.
\newblock In {\em NIPS}, 2011.

\bibitem{olshausen1997sparse}
B.~A. Olshausen and D.~J. Field.
\newblock Sparse coding with an overcomplete basis set: A strategy employed by
  v1?
\newblock {\em Vision research}, 1997.

\bibitem{partalas2015lshtc}
I.~Partalas, A.~Kosmopoulos, N.~Baskiotis, T.~Artieres, G.~Paliouras,
  E.~Gaussier, I.~Androutsopoulos, M.-R. Amini, and P.~Galinari.
\newblock Lshtc: A benchmark for large-scale text classification.
\newblock {\em arXiv:1503.08581 [cs.IR]}, 2015.

\bibitem{roller2004max}
B.~T. C. G.~D. Roller.
\newblock Max-margin markov networks.
\newblock {\em Advances in neural information processing systems}, 16:25, 2004.

\bibitem{feature_cluster}
C.~Scherrer, A.~Tewari, M.~Halappanavar, and D.~Haglin.
\newblock Feature clustering for accelerating parallel coordinate descent.
\newblock {\em NIPS}, 2012.

\bibitem{shalev2013stochastic}
S.~Shalev-Shwartz and T.~Zhang.
\newblock Stochastic dual coordinate ascent methods for regularized loss.
\newblock {\em JMLR}, 2013.

\bibitem{teh2006hierarchical}
Y.~W. Teh, M.~I. Jordan, M.~J. Beal, and D.~M. Blei.
\newblock Hierarchical dirichlet processes.
\newblock {\em Journal of the american statistical association}, 101(476),
  2006.

\bibitem{terry2013replicated}
D.~Terry.
\newblock Replicated data consistency explained through baseball.
\newblock {\em CACM}, 2013.

\bibitem{tibshirani1996regression}
R.~Tibshirani.
\newblock Regression shrinkage and selection via the lasso.
\newblock {\em Journal of the Royal Statistical Society. Series B
  (Methodological)}, pages 267--288, 1996.

\bibitem{valiant1990bridging}
L.~G. Valiant.
\newblock A bridging model for parallel computation.
\newblock {\em Communications of the ACM}, 33(8):103--111, 1990.

\bibitem{ventures2006stanley}
M.~D. Ventures.
\newblock Stanley: The robot that won the darpa grand challenge.
\newblock {\em Journal of field Robotics}, 23(9):661--692, 2006.

\bibitem{wainwright2008graphical}
M.~J. Wainwright and M.~I. Jordan.
\newblock Graphical models, exponential families, and variational inference.
\newblock {\em Foundations and Trends{\textregistered} in Machine Learning},
  1(1-2):1--305, 2008.

\bibitem{wei2015managed}
J.~Wei, W.~Dai, A.~Qiao, Q.~Ho, H.~Cui, G.~R. Ganger, P.~B. Gibbons, G.~A.
  Gibson, and E.~P. Xing.
\newblock Managed communication and consistency for fast data-parallel
  iterative analytics.
\newblock In {\em Proceedings of the Sixth ACM Symposium on Cloud Computing},
  pages 381--394. ACM, 2015.

\bibitem{xie2015distributed}
P.~Xie, J.~K. Kim, Y.~Zhou, Q.~Ho, A.~Kumar, Y.~Yu, and E.~Xing.
\newblock Distributed machine learning via sufficient factor broadcasting.
\newblock {\em arXiv preprint arXiv:1511.08486}, 2015.

\bibitem{xing2015petuum}
E.~Xing, Q.~Ho, W.~Dai, J.-K. Kim, J.~Wei, S.~Lee, X.~Zheng, P.~Xie, A.~Kumar,
  and Y.~Yu.
\newblock Petuum: A new platform for distributed machine learning on big data.
\newblock {\em Big Data, IEEE Transactions on}, PP(99):1--1, 2015.

\bibitem{cmu10708}
E.~P. Xing.
\newblock Graphical models class.
\newblock \url{http://www.cs.cmu.edu/~epxing/Class/10708/index.html}, 2014.

\bibitem{xing2002distance}
E.~P. Xing, M.~I. Jordan, S.~Russell, and A.~Y. Ng.
\newblock Distance metric learning with application to clustering with
  side-information.
\newblock In {\em NIPS}, 2002.

\bibitem{yang2013trading}
T.~Yang.
\newblock Trading computation for communication: Distributed stochastic dual
  coordinate ascent.
\newblock In {\em NIPS}, 2013.

\bibitem{yao2009efficient}
L.~Yao, D.~Mimno, and A.~McCallum.
\newblock Efficient methods for topic model inference on streaming document
  collections.
\newblock In {\em Proceedings of the 15th ACM SIGKDD international conference
  on Knowledge discovery and data mining}, pages 937--946. ACM, 2009.

\bibitem{yuan2015lightlda}
J.~Yuan, F.~Gao, Q.~Ho, W.~Dai, J.~Wei, X.~Zheng, E.~P. Xing, T.-Y. Liu, and
  W.-Y. Ma.
\newblock Lightlda: Big topic models on modest compute clusters.
\newblock In {\em WWW}. 2015.

\bibitem{yuan2006model}
M.~Yuan and Y.~Lin.
\newblock Model selection and estimation in regression with grouped variables.
\newblock {\em Journal of the Royal Statistical Society: Series B (Statistical
  Methodology)}, 2006.

\bibitem{zaharia2012resilient}
M.~Zaharia, M.~Chowdhury, T.~Das, A.~Dave, J.~Ma, M.~McCauley, M.~J. Franklin,
  S.~Shenker, and I.~Stoica.
\newblock Resilient distributed datasets: A fault-tolerant abstraction for
  in-memory cluster computing.
\newblock In {\em NSDI}, 2012.

\bibitem{zhang2015poseidon}
H.~Zhang, Z.~Hu, J.~Wei, P.~Xie, G.~Kim, Q.~Ho, and E.~Xing.
\newblock Poseidon: A system architecture for efficient gpu-based deep learning
  on multiple machines.
\newblock {\em arXiv preprint arXiv:1512.06216}, 2015.

\bibitem{zhang2004solving}
T.~Zhang.
\newblock Solving large scale linear prediction problems using stochastic
  gradient descent algorithms.
\newblock In {\em Proceedings of the twenty-first international conference on
  Machine learning}, page 116. ACM, 2004.

\bibitem{zhao2014quasi}
B.~Zhao and E.~P. Xing.
\newblock Quasi real-time summarization for consumer videos.
\newblock In {\em Computer Vision and Pattern Recognition (CVPR), 2014 IEEE
  Conference on}, pages 2513--2520. IEEE, 2014.

\bibitem{zheng2015model}
X.~Zheng, J.-K. Kim, Q.~Ho, and E.~Xing.
\newblock Model-parallel inference for big topic models.
\newblock {\em To appear in Big Data, IEEE Transactions on}, 2015.

\bibitem{mspg}
Y.~Zhou, Y.~Yu, W.~Dai, Y.~Liang, and E.~P. Xing.
\newblock On convergence of model parallel proximal gradient algorithm for
  stale synchronous parallel system.
\newblock In {\em The $19^{th}$ International Conference on Artificial
  Intelligence and Statistics {(AISTATS)}}, 2016.

\bibitem{zhu2009medlda}
J.~Zhu, A.~Ahmed, and E.~P. Xing.
\newblock Medlda: maximum margin supervised topic models for regression and
  classification.
\newblock In {\em Proceedings of the 26th annual international conference on
  machine learning}, pages 1257--1264. ACM, 2009.

\bibitem{zhu2014bayesian}
J.~Zhu, N.~Chen, and E.~P. Xing.
\newblock Bayesian inference with posterior regularization and applications to
  infinite latent svms.
\newblock {\em The Journal of Machine Learning Research}, 15(1):1799--1847,
  2014.

\bibitem{zhu2009maximum}
J.~Zhu and E.~P. Xing.
\newblock Maximum entropy discrimination markov networks.
\newblock {\em The Journal of Machine Learning Research}, 10:2531--2569, 2009.

\end{thebibliography}
